\documentclass{article}

    \PassOptionsToPackage{super, sort&compress, comma}{natbib}
\usepackage[preprint]{neurips_2025}

\usepackage[utf8]{inputenc} % allow utf-8 input
\usepackage[T1]{fontenc}    % use 8-bit T1 fonts
\usepackage{hyperref}       % hyperlinks
\usepackage{url}            % simple URL typesetting
\usepackage{booktabs}       % professional-quality tables
\usepackage{amsfonts}       % blackboard math symbols
\usepackage{nicefrac}       % compact symbols for 1/2, etc.
\usepackage{microtype}      % microtypography

\usepackage{titlesec}  % (in preamble, not after \appendix)
\titleformat{\section}
  {\normalfont\Large\bfseries}{SI\,\thesection}{1em}{}
\titleformat{\subsection}
  {\normalfont\large\bfseries}{SI\,\thesubsection}{1em}{}
\titleformat{\subsubsection}
  {\normalfont\normalsize\bfseries}{SI\,\thesubsubsection}{1em}{}

\usepackage{amsmath,amsfonts,bm}

\def\1{\bm{1}}

\def\rvs{{\mathbf{s}}}

\def\rvw{{\mathbf{w}}}
\def\rvx{{\mathbf{x}}}

\DeclareMathAlphabet{\mathsfit}{\encodingdefault}{\sfdefault}{m}{sl}
\SetMathAlphabet{\mathsfit}{bold}{\encodingdefault}{\sfdefault}{bx}{n}

\def\gD{{\mathcal{D}}}

\def\gW{{\mathcal{W}}}
\def\gX{{\mathcal{X}}}

\usepackage{algorithm}
\usepackage{algpseudocode}
\usepackage{tabularx}
\usepackage[dvipsnames]{xcolor}
\usepackage{nicefrac}
\usepackage{xurl}
\usepackage{enumitem}
\usepackage{authblk}
\usepackage[capitalize,noabbrev]{cleveref}
\usepackage[breakable,skins]{tcolorbox}
\usepackage[acronym]{glossaries}
\glsdisablehyper

\hypersetup{
  colorlinks,
  linkcolor={blue!50!black},
  citecolor={blue!50!black},
  urlcolor={blue!50!black}
}

\newcommand{\sx}[1]{{\color{teal} [\textbf{SX:} #1]}}

\newacronym{bo}{BO}{Bayesian optimization}
\newacronym{pi}{PI}{probability of improvement}
\newacronym{ucb}{UCB}{upper confidence bound}
\newacronym{pv}{PV}{posterior variance}
\newacronym{hte}{HTE}{high-throughput experimentation}
\newacronym{gp}{GP}{Gaussian process}
\newacronym{ei}{EI}{expected improvement}
\newacronym{kg}{KG}{knowledge gradient}
\newacronym{ts}{TS}{Thompson Sampling}
\newacronym{saa}{SAA}{sample-average approximation}

\newglossaryentry{currybo}{name={\textsc{CurryBO}},description={}}
\newglossaryentry{seqonelaucbpv}{name={\texttt{Seq 1LA-UCB-PV}},description={}}
\newglossaryentry{seqonelaucbzeroonepv}{name={\texttt{Seq 1LA-UCB(0.1)-PV}},description={}}
\newglossaryentry{seqonelaucbzerofivepv}{name={\texttt{Seq 1LA-UCB(0.5)-PV}},description={}}
\newglossaryentry{seqtwolaucbfivepv}{name={\texttt{Seq 2LA-UCB(5.0)-PV}},description={}}
\newglossaryentry{seqonelapipv}{name={\texttt{Seq 1LA-PI-PV}},description={}}
\newglossaryentry{seqonelaucbra}{name={\texttt{Seq 1LA-UCB-RA}},description={}}
\newglossaryentry{seqtwolaucbpv}{name={\texttt{Seq 2LA-UCB-PV}},description={}}
\newglossaryentry{seqtwolaucbra}{name={\texttt{Seq 2LA-UCB-RA}},description={}}
\newglossaryentry{jointtwolaucb}{name={\texttt{Joint 2LA-UCB}},description={}}
\newglossaryentry{jointtwolaei}{name={\texttt{Joint 2LA-EI}},description={}}
\newglossaryentry{seqonelaucbsingle}{name={\texttt{Seq 1LA-UCB-Single}},description={}}
\newglossaryentry{seqonela}{name={\texttt{Seq 1LA}},description={}}
\newglossaryentry{seqtwola}{name={\texttt{Seq 2LA}},description={}}
\newglossaryentry{jointtwola}{name={\texttt{Joint 2LA}},description={}}
\newglossaryentry{bandit}{name={\texttt{bandit}},description={}}
\newglossaryentry{random}{name={\texttt{random}},description={}}

\newglossaryentry{ucbfunc}{name={\texttt{UCB}},description={}}
\newglossaryentry{ucbfuncs}{name={\texttt{UCB(s)}},description={}}
\newglossaryentry{pvfunc}{name={\texttt{PV}},description={}}
\newglossaryentry{pifunc}{name={\texttt{PI}},description={}}
\newglossaryentry{eifunc}{name={\texttt{EI}},description={}}
\newglossaryentry{rafunc}{name={\texttt{RA}},description={}}
\newglossaryentry{singlefunc}{name={\texttt{SINGLE}},description={}}

\newglossaryentry{mean}{name={\texttt{mean}},description={}}
\newglossaryentry{threshold}{name={\texttt{threshold}},description={}}
\newglossaryentry{gap}{name={\texttt{GAP}},description={}}

\title{Bayesian Optimization for General Reaction Conditions}
\author[a,b]{Stefan P.~Schmid\thanks{Correspondence to: \texttt{stefan.schmid@chem.ethz.ch}, \texttt{aspuru@utoronto.ca}, \texttt{akristi@uwo.ca}, \texttt{kjell.jorner@chem.ethz.ch}, \texttt{strieth-kalthoff@uni-wuppertal.de}.}$\;^{,}$}
\author[c,d]{Ella Miray Rajaonson}
\author[c,d]{Cher Tian Ser}
\author[e,f]{Mohammad Haddadnia}
\author[c,g]{Shi Xuan Leong}
\author[c,d,h,i,j,k,l,m,n]{Alán Aspuru-Guzik\textsuperscript{$\ast,$}}
\author[d,o]{Agustinus Kristiadi\textsuperscript{$\ast,$}}
\author[a,b]{Kjell Jorner\textsuperscript{$\ast,$}}
\author[p]{Felix Strieth-Kalthoff\textsuperscript{$\ast,$}}
\affil[a]{Institute of Chemical and Bioengineering, Department of Chemistry and Applied Biosciences, ETH Zurich, Zurich CH-8093, Switzerland}
\affil[b]{NCCR Catalysis, Switzerland}
\affil[c]{Department of Chemistry, University of Toronto, Toronto, Canada}
\affil[d]{Vector Institute, Toronto, Canada}
\affil[e]{Department of Biological Chemistry \& Molecular Pharmacology, Harvard Medical School, Boston, MA, USA}
\affil[f]{Dana-Farber Cancer Institute, Boston, MA, USA}
\affil[g]{School of Chemistry, Chemical Engineering and Biotechnology, Nanyang Technological University, Singapore}
\affil[h]{Department of Computer Science, University of Toronto, Toronto, Canada}
\affil[i]{Department of Chemical Engineering and Applied Chemistry, University of Toronto, Toronto, Canada}
\affil[j]{Department of Materials Science and Engineering, University of Toronto, Toronto, Canada}
\affil[k]{Acceleration Consortium, University of Toronto, Toronto, Canada}
\affil[l]{Canadian Institute for Advanced Research (CIFAR)}
\affil[m]{Institute of Medical Science, Medical Sciences Building, Toronto, Canada}
\affil[n]{NVIDIA, Toronto, Canada}
\affil[o]{Department of Computer Science, Western University, London, Canada}
\affil[p]{School of Mathematics and Natural Sciences, University of Wuppertal, Wuppertal, Germany}

\begin{document}

\maketitle
\vspace{-0.75cm}

\begin{abstract}
\vspace{-0.2cm}
General chemical reaction conditions that achieve consistently high performance across multiple substrates are important for practical applications such as library synthesis and high-throughput experimentation. 
However, identifying such conditions efficiently has been a longstanding challenge, as it requires decision making under uncertainty with respect to both conditions and substrates, while minimizing the number of required experiments. 
Here, we introduce \gls{currybo}, a high-level framework for generality-oriented optimization.
By formalizing the problem as Bayesian optimization over curried functions, \gls{currybo} provides a unified framework that accommodates different generality definitions (\textit{e.g.} mean yield across substrates), and supports a range of substrate and condition selection strategies.
We evaluate this framework on four benchmark tasks in experimental reaction optimization, and systematically analyze key algorithmic components. 
Our results show that efficient experiment planning can be achieved by emphasizing exploration when selecting reaction conditions, followed by the uncertainty-guided prioritization of substrates in a sequential decison-making scheme.
Based on these insights, we design and validate an optimization policy that substantially improves sample efficiency relative to previously reported approaches across all benchmarks. 
Overall, the flexibility and modularity of \gls{currybo} facilitate the integration of generality-oriented optimization into experimental settings, enabling more efficient identification of solutions that perform robustly across diverse tasks. 
\end{abstract}

\glsresetall{}
\renewcommand{\thefootnote}{\fnsymbol{footnote}}

%!TEX root=../main.tex

General reaction conditions, \textit{i.e.}, conditions that consistently yield high reaction outcomes across a range of related molecules, are a central goal for synthesis in the pharmaceutical and chemical industries \citep{wagen_screening_2022, prieto_kullmer_accelerating_2022, rein_generality-oriented_2023, betinol_data-driven_2023, rana_standardizing_2024, schmid_catalysing_2024, gallarati_genetic_2024, zacate_considerations_2025}.
Once identified, such general conditions enable efficient synthesis of compound libraries for experimental screening \citep{dombrowski_chosen_2022, angello_closed-loop_2022, kowalski_automated_2023, strieth-kalthoff_delocalized_2024, husbands_high-throughput_2025}, parallel handling of substrates in \gls{hte} settings \citep{dombrowski_chosen_2022, angello_closed-loop_2022, strieth-kalthoff_delocalized_2024}, or a broad substrate scope for developing novel synthetic techniques \citep{wagen_screening_2022, prieto_kullmer_accelerating_2022, rein_generality-oriented_2023, zacate_considerations_2025}.

\begin{figure*}[th]
    \centering
    \includegraphics[]{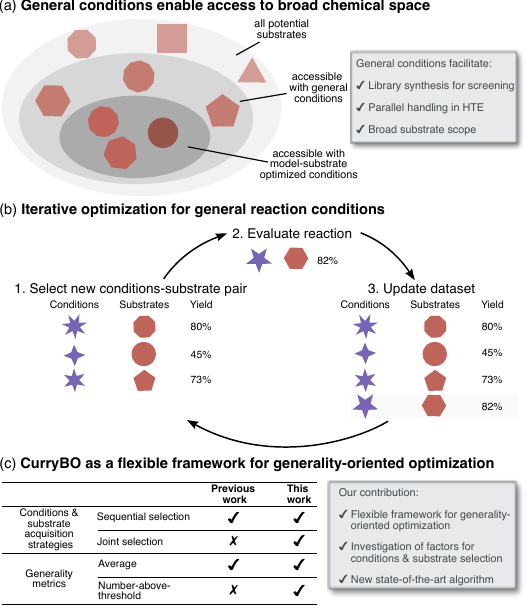}
    %\vspace{-1em}
    \caption{
        (a) Conditions identified via single-substrate optimization typically limit the viable substrate space relative to general conditions. Such general conditions facilitate library synthesis, HTE handling, and a broad substrate scope. (b) Generality-oriented optimization considers multiple substrates, and iteratively selects conditions and substrates for experiments, which are performed and used to update the dataset. (c) \gls{currybo}, a flexible framework for generality-oriented optimization integrates custom strategies for conditions and substrate selection, as well as generality metrics. Systematic evaluation of factors influencing optimization efficiency enables the design of a new state-of-the-art algorithm for generality-oriented optimization.
    }
    \label{fig:generality-oriented-optimization}
\end{figure*}

In contrast, current practice typically performs reaction optimization on a single model substrate, after which the resulting conditions are applied to all other substrates of interest \citep{wagen_screening_2022}. 
This traditionally human-centric process can be accelerated by iterative, model-based techniques such as \gls{bo} \citep{clayton_algorithms_2019, shields_bayesian_2021, guo_bayesian_2023}, which are increasingly used in autonomous "self-driving" laboratories \citep{tom_self-driving_2024}. 
However, optimization on a single substrate does not ensure generality: the identified conditions are not guaranteed or even likely to be applicable to unseen substrates (see \cref{fig:generality-oriented-optimization}a) \citep{wagen_screening_2022, wang_identifying_2024}. 

Translating this framework to the optimization of general conditions is associated with a considerably higher cost. 
In principle, assessing the generality of a set of reaction conditions requires evaluating them across multiple substrates, leading to a combinatorial growth in the number of experiments.
Reducing this cost necessitates algorithms that can (a) estimate the generality of a set of conditions from sparse experiments on a subset of substrates, and (b) jointly propose the next set of conditions and substrates to evaluate next. 
These requirements go beyond standard \gls{bo} formulations, and require specialized adaptations. 

Recent studies have taken important steps in this direction. 
In a pioneering study, Grzybowski, Burke and co-workers optimized conditions for a Suzuki–Miyaura coupling by maximizing the average yield across substrates using a BO-based approach \citep{angello_closed-loop_2022}.
Later, Doyle and co-workers formulated the problem as a multi-armed bandit over discrete sets of reaction conditions \citep{wang_identifying_2024}.
Both studies highlight the practical utility of generality-oriented optimization, but each adopts specific modeling and decision-making choices. 
In particular, both works (i) decouple condition and substrate selection by first selecting promising conditions and subsequently identifying substrates for evaluation \citep{toscano-palmerin_bayesian_2022}, (ii) define generality as average yield, neglecting other practically relevant generality metrics \citep{betinol_data-driven_2023}, and (iii) do not systematically investigate exploration-exploitation tradeoffs in condition and substrate selection.
As a result, the broader algorithmic design space for generality-oriented optimization (\cref{fig:generality-oriented-optimization}c) remains underexplored, leaving efficient algorithms undiscovered and delaying robust implementation in (chemical) laboratories.

In this work, we address this gap by introducing \gls{currybo}, a flexible framework for generality-oriented Bayesian optimization that enables systematic exploration of the algorithmic design space (see \cref{fig:generality-oriented-optimization}c).
We formalize generality-oriented optimization as an optimization problem over \textit{curried} functions \citep{frege_grundgesetze_1893} \footnote{By currying, we refer to the mathematical formalism of translating functions with multiple arguments (in our case conditions and substrates) into a family of functions with one argument each (conditions).}.
This formalization enables the flexible incorporation of (a) customizable generality metrics, and (b) different strategies for condition and substrate selection, including both sequential and joint selection, as well as different exploration-exploitation trade-offs.
This formulation provides a unified basis for benchmarking algorithmic choices.

Using the \gls{currybo} framework, we investigate four key questions:  
(1) Does optimization using multiple substrates actually lead to more transferable conditions relative to single-substrate approaches? 
(2) Does joint conditions and substrate selection improve optimization efficiency? 
(3) How does the exploration–exploitation tradeoff in substrate and condition acquisition affect optimization behavior? 
(4) Can we use the insights to identify a strategy for generality-oriented optimization that improves over the current state of the art? 
These aspects are evaluated across several experimental datasets of multi-substrate reaction optimization. 
Our results clearly confirm that the incorporation of multiple substrates during optimization improves the transferability of identified conditions. 
Efficient optimization can be achieved using a highly-explorative sequential strategy that first explores condition space and subsequently selects substrates based on predictive uncertainty, while joint acquisition does not provide a consistent advantage.
Based on these insights, we identify and validate an optimization policy that consistently improves over state-of-the-art, highlighting the importance of empirical benchmarking for practical optimization.
More broadly, this work lays the foundation for the integration of generality-oriented optimization in chemical laboratories, ultimately yielding more robust reactions that can be applied towards successful chemical library synthesis, synthetic method development, and as starting points for specialized reaction optimizations.
%!TEX root=../main.tex

\section*{Results}

\subsubsection*{Problem formulation: Optimizing for generality in curried function space}

\begin{figure*}[t]
    \centering
    \includegraphics[width=\linewidth]{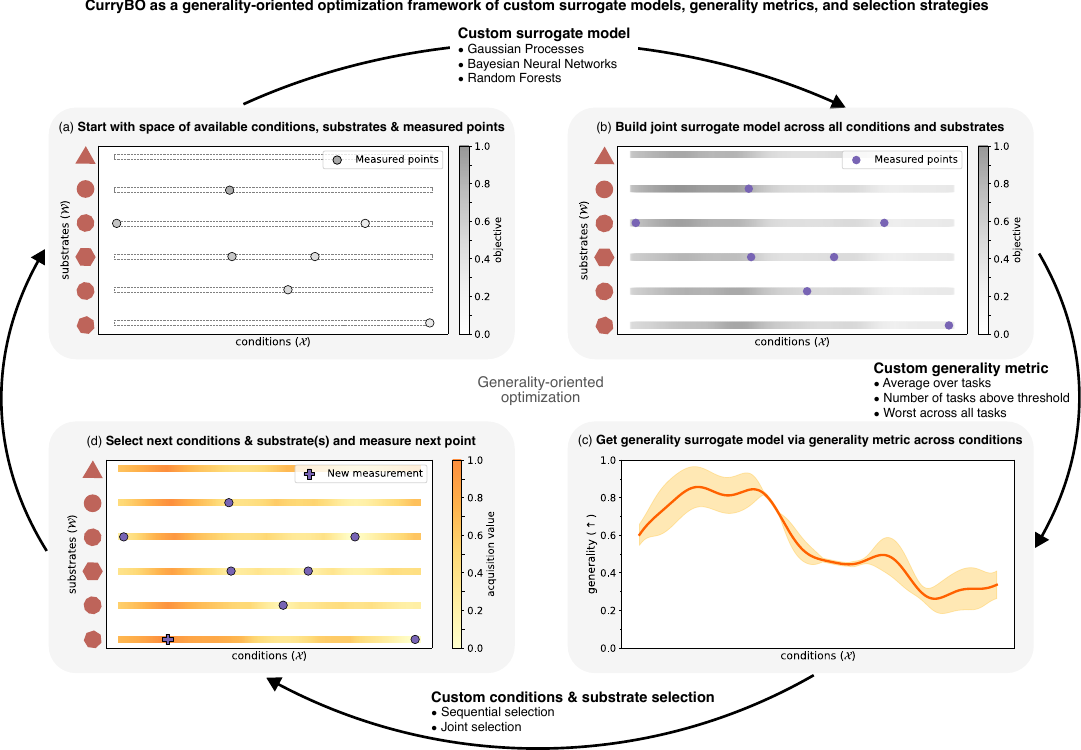}
    \vspace{-1em}
    \caption{
        Workflow of generality-oriented optimization. Starting from measurements across multiple parameters and discrete tasks (a), a surrogate model over the joint space is trained (b). Using a generality metric, the surrogate model is used to obtain a posterior over the generality metric (c). Given an acquisition strategy for selecting the next parameters and task, the selected point is measured and the next iteration commences (d). CurryBO serves as a framework to combine custom surrogate models, generality metrics and acquisition strategies.
    }
    \label{fig:partial-monitoring-conceptual}
\end{figure*}

To derive the central formalism in \gls{currybo}, we begin by formalizing the problem of generality-oriented optimization.
We consider a black-box function $f: \gX \times \gW \rightarrow \mathbb{R}$ defined over the joint space $\gX \times \gW$.
Here, $\rvx \in \gX$ denotes continuous, discrete or mixed-variable parameters (\textit{i.e.}, conditions) and $\gW = \{ \rvw_i \}_{i=1}^n$ is a $n$-sized discrete space of tasks (\textit{i.e.}, substrates).
The function output is a real number (\textit{e.g.} reaction yield), as shown in \cref{fig:partial-monitoring-conceptual}a.
Currying over the second argument of $f(\rvx, \rvw)$ produces a family of substrate-specific functions $f_{\rvw}: \gX \rightarrow \mathbb{R}$ that can be evaluated experimentally.
In the chemical reaction setting, evaluating $f_{\rvw_i}(\rvx)$ corresponds to measuring the reaction outcome of a substrate (parametrized by $\rvw_\text{i}$) under reaction conditions $\rvx$.
Applying a generality metric (\textit{e.g.}, the mean) over all members of this function family defines a generality function $\phi(\rvx)$, which is the eventual optimization objective.

This curried formulation separates the problem into two distinct components: (i) modeling the family of functions $\{f_{\rvw}\}$ over $\gX$, and (ii) aggregating their values into a generality objective.
Measuring the generality of a set of conditions $\rvx$ exactly requires evaluating $f_{\rvw_i}$ for all $n$ substrates.
For practically relevant values of $n$, this requires an infeasible number of experiments.
In practice, generality can therefore only be \emph{partially monitored} \citep{rustichini_minimizing_1999, lattimore_cleaning_2019, lattimore_bandit_2020}, \textit{i.e.}, each set of conditions is evaluated on only a subset of substrates.
This setting introduces two important distinctions from standard \gls{bo}:
(i) The acquisition strategy must propose both the next conditions \textit{and} the substrate to evaluate.
(ii) After $k$ experiments, the optimal set of conditions cannot be directly identified from all previous observations, as each set of conditions has only been partially evaluated across substrates.
Instead, the optimum $\hat{\rvx} = \text{argmax}_{\rvx \in \gX} \ \phi(\rvx)$ must be inferred indirectly from a surrogate model of the generality function $\phi(\rvx)$, trained on all available observations.
Therefore, the resulting optimization problem can be stated as follows: 
Given a budget of $K$ experiments, iteratively recommend conditions--substrate pairs to maximize $\phi(\hat{\rvx}_{K})$, where $\hat{\rvx}_{K}$ is the optimal set of conditions after $K$ experiments.
Further details are outlined in the Methods section.

To address this setting, we developed the \gls{currybo} framework, which leverages the curried structure of the problem, and enables flexible and modular choices of all optimizer components: surrogate model, generality metrics, and acquisition policies.
At each iteration, \gls{currybo} fits a probabilistic surrogate model $g(\rvx, \rvw)$ over the joint $\gX \times \gW$ space using all available $k$ observations (\cref{fig:partial-monitoring-conceptual}b).
From the posterior of $g$, the posterior over the generality function $\phi(\rvx)$ can be estimated \textit{via} Monte-Carlo integration for any functional form of the generality metric (\cref{fig:partial-monitoring-conceptual}c). 
Based on this generality posterior, an \textit{acquisition strategy} selects the next experiment (\cref{fig:partial-monitoring-conceptual}d), \textit{i.e.} the next conditions--substrate pair ($\rvx_{k+1},\rvw_{k+1}$) \footnote{In this work, an \textit{acquisition strategy} specifies how candidate points are selected from the joint space $\gX \times \gW$ via the tractable optimization of associated \textit{acquisition functions}.}.
The measurement is added to the dataset, initiating a new iteration.

Formally, selecting ($\rvx_{k+1},\rvw_{k+1}$) corresponds to a lookahead acquisition problem over the joint $\gX \times \gW$ space. 
In other words, the acquisition strategy targets decision quality by selecting the experiment that maximizes the expected generality of the ultimately recommended optimum under the updated model, \textit{i.e.}, that maximizes $\phi(\hat{\rvx}_{k+1})$.
In \gls{currybo}, we consider two classes of acquisition strategies.
\begin{itemize}
\item \textit{Joint} selection of ($\rvx_{k+1},\rvw_{k+1}$) by maximizing a two-step lookahead acquisition function over $\gX \times \gW$ \citep{toscano-palmerin_bayesian_2022}.
\item \textit{Sequential} recommendation of $\rvx_{k+1}$ and $\rvw_{k+1}$, following literature precendents \citep{angello_closed-loop_2022, wang_identifying_2024}. This approximation decomposes the acquisition over $\gX \times \gW$ into two sequential optimizations of classical acquisition functions, first over $\gX$ and then over $\gW$.
\end{itemize}
The specific choice of acquisition function controls the exploration–exploitation tradeoff in each step. A formal treatment of these strategies and their specific implementations are provided in the Supplementary Information \cref{App:acqf}.
.
%In this study, we investigate sequential one-step lookahead (\gls{seqonela}), sequential two-step lookahead (\gls{seqtwola}) and joint two-step lookahead (\gls{jointtwola}) acquisition strategies.
%Note that one-step lookahead refers to ``classical'' acquisition functions (\textit{e.g.}, \gls{ucbfunc}).

\subsubsection*{Upper-bound analysis: Multi-substrate optimization yields more transferable optima}

At the outset of our investigations, we examine the commonly assumed \citep{wagen_screening_2022,angello_closed-loop_2022, wang_identifying_2024} but thus far untested hypothesis that generality-oriented optimization using multiple substrates actually leads to more general reaction conditions.
To test this hypothesis, we use four experimental chemical reaction datasets from the literature: (1) Pd-catalyzed carbon-heteroatom coupling \citep{buitrago_santanilla_nanomole-scale_2015}, (2) asymmetric \textit{N},\textit{S}-acetal formation \citep{zahrt_prediction_2019}, (3) borylation \citep{stevens_advancing_2022}, and (4) deoxyfluorination \citep{nielsen_deoxyfluorination_2018}.
Each dataset consists of multiple substrates evaluated under up to 96 reaction conditions in an \gls{hte} setting, spanning a combinatorial search space.
To mitigate the known bias of \gls{hte} datasets toward high-outcome experiments\citep{strieth-kalthoff_machine_2022, beker_machine_2022}, we augment the datasets with low-outcome examples using a chemically sensible search-space expansion workflow (see Supporting Information \cref{subsubsec: augmentation}).

We assess the transferability of general optima as follows: (1) substrates are randomly divided into train and test sets; (2) the most general conditions on the training data are identified through exhaustive search for varying train set sizes; and (3) these conditions are evaluated on the withheld test set.
Transferability to test substrates is quantified using a normalized generality score (see Methods), evaluated for two generality metrics: (1) the average outcome across substrates (mean generality metric, \gls{mean}) and (2) the number of substrates exceeding a predefined outcome threshold (threshold generality metric, \gls{threshold}).
Practically, this analysis provides an upper bound on what BO-based strategies can achieve.

\begin{figure*}[t]
    \centering
    \includegraphics[width=\linewidth]{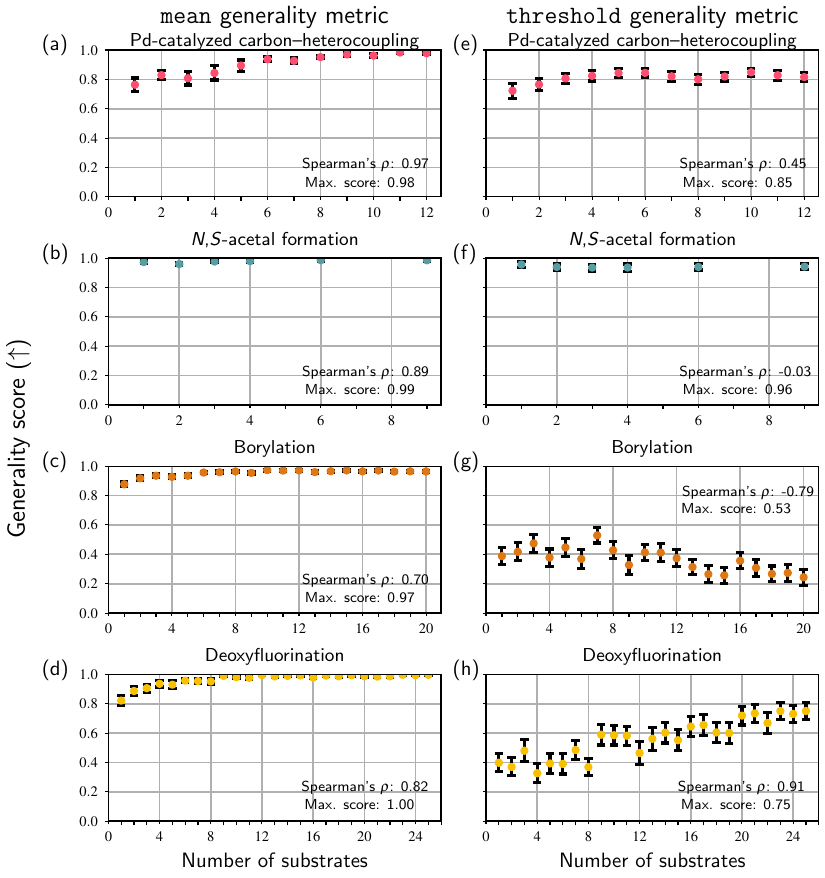}
    \vspace{-1em}
    \caption{
        Test-set generality score given by the \gls{mean} (a)-(d) and \gls{threshold} (e)-(h) metrics, as determined by exhaustive grid search for the four datasets. Average and standard error are calculated from the 30 different train/test substrate splits.
    }
    \label{fig:grid-search}
\end{figure*}

For the \gls{mean} generality metric, we observe that the maximum achievable transferability increases with the number of training substrates (\cref{fig:grid-search}, a--d).
Transferability improves significantly as the number of training substrates increases up to approximately $n\sim6$, after which gains diminish.
The only exception is the \textit{N},\textit{S}-acetal formation dataset, where near-maximal transferability is already achieved with a single train substrate (\cref{fig:grid-search}b).
This behaviour may reflect strong catalyst control, leading to consistent performance across substrates\citep{betinol_controlling_2025}, such that single-substrate optimization already yields general conditions.

The \gls{threshold} metric shows less consistent behavior across reactions (\cref{fig:grid-search}, e--h).
As above, the \textit{N},\textit{S}-acetal formation shows high generality with a single train substrate.
However, among the remaining benchmark problems, only two show a clear increase in transferability with an increasing number of training substrates (shown by the Spearman's $\rho$). 
For the borylation and deoxyfluorination reactions, transferability is lower than for the \gls{mean} generality metric, and the improvements with additional train substrates are less pronounced. 
Careful analysis of the underlying yield distributions indicates that reaction yields are highly sensitive to specific conditions--substrate combinations (see Supporting Information \cref{subsubsec:dataset_details}). 
Because the \gls{threshold} metric assigns identical scores to yields just below the threshold (90$\%$) as substantially lower yields, it discrards information about near-optimal performance, thereby reducing its ability to distinguish broadly effective conditions.
In the borylation dataset, several sets of conditions achieve identical optimal \gls{threshold} generality on the full dataset by performing well on disjoint sets of substrates.
As a result, grid search becomes highly sensitive to the specific train/test split: larger training sets may overrepresent specific substrate subsets, leading to overfitting and reduced transferability on the test set.
In contrast, in datasets where some conditions perform consistently well across substrates, the signal for generality is stronger, and the \gls{threshold} metric remains informative.
Overall, although the magnitude of the effect is reaction-dependent, these results support the hypothesis that multi-substrate optimization leads to more general conditions \citep{wagen_screening_2022, angello_closed-loop_2022, wang_identifying_2024}, and underline the utility of generality-oriented optimization.

\subsubsection*{Acquisition strategies:  sequential condition–substrate acquisition is sufficient}

Having established the utility of generality-oriented optimization, we next investigate which factors enable sample-efficient optimization in this scenario.
First, we examine the effect of decoupling condition and substrate selection by comparing joint and sequential acquisition strategies.
Specificially, we evaluate the following acquisition strategies: 
(i) joint selection of ($\rvx_{k+1}$, $\rvw_{k+1}$) by maximizing a two-step lookahead acquisition function over $\gX \times \gW$ jointly (\gls{jointtwola}); 
(ii) sequential selection of $\rvx_{k+1}$ and $\rvw_{k+1}$ by maximizing two-step lookahead acquisition functions over $\gX$ and $\gW$ sequentially (\gls{seqtwola}); and 
(iii) sequential acquisition by maximizing two one-step lookahead acquisition functions first over $\gX$ and then over $\gW$ (\gls{seqonela}).
For both sequential strategies, we use the Upper Confidence Bound (\gls{ucbfunc}) for condition acquisition (\textit{i.e.}, $\rvx_{k+1}$), and the Posterior Variance (\gls{pvfunc}) for substrate acquisition (\textit{i.e.}, $\rvw_{k+1}$) \citep{angello_closed-loop_2022}. 
\gls{jointtwola} is performed using the \gls{ucbfunc} acquisition function.

For each benchmark problem, we perform 30 independent optimization runs, differing in their randomly sampled initial observations and the substrates considered for optimization.
To enable comparison across datasets, we report the \gls{gap} value\citep{jiang_binoculars_2020}, defined as $\text{\gls{gap}} = (y_k - y_0) / (y^* - y_0)$. 
Here, $y_k$ and $y_0$ denote the generality values $\phi(\hat{\rvx}_{k})$ and $\phi(\hat{\rvx}_{0})$ after $k$ experiments and the initialization experiment $0$, respectively, and $y^*$ is the maximum achievable generality for the dataset. 
Thus, the \gls{gap} represents normalized performance between $0$ and $1$, where $\text{\gls{gap}} = 1$ corresponds to the optimum.
We report averages across all four benchmark problems, along  with standard errors of the mean for uncertainty estimation. Problem-specific optimization trajectories are provided in the Supplementary Information (\cref{subsubsec:add_results_augmented}).

\begin{figure*}[t]
    \centering
    \includegraphics[width=\linewidth]{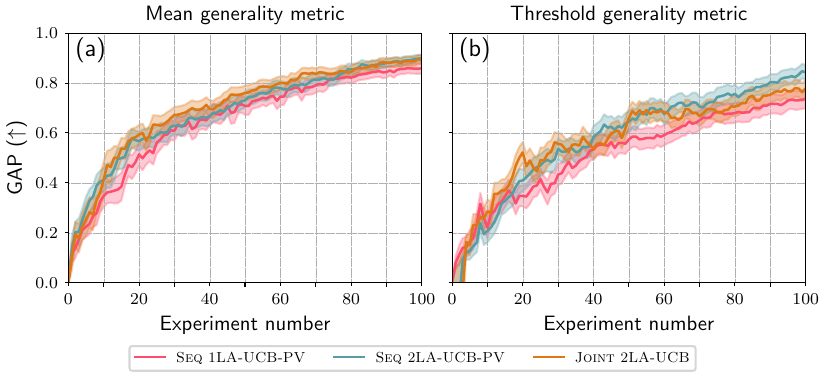}
    \vspace{-1em}
    \caption{
        Optimization trajectories, averaged over all datasets, for the sequential one-step lookahead (\gls{seqonelaucbpv}), sequential two-step lookahead (\gls{seqtwolaucbpv}) and joint two-step lookahead (\gls{jointtwolaucb}) acquisition strategies on the \gls{mean} (a) and \gls{threshold} (b) generality metrics.
    }
    \label{fig:strategies}
\end{figure*}

The optimization trajectories show that sequential conditions--substrate acquisition achieves similar (\gls{mean}, \cref{fig:strategies}a) or slightly increased (\gls{threshold}, \cref{fig:strategies}b) \gls{gap} values compared to a joint acquisition.
This result indicates that, empirically, decoupling condition and substrate selection is a valid assumption for chemical reaction optimization that does not degrade performance.
As sequential acquisition places a higher emphasis on condition selection, these observations suggest that optimization performance is primarily governed by condition selection, and that jointly selecting specific conditions--substrate combinations does not provide notable sample efficiency gains.

Comparing the sequential acquisition strategies, the two-step lookahead approach achieves similar (\gls{mean}, \cref{fig:strategies}a) or slightly improved (\gls{threshold}, \cref{fig:strategies}b) \gls{gap} values relative to the one-step lookahead strategy.
Notable differences are only observed for the \gls{threshold} generality metric, and emerge primarily at later stages of the optimization.
Analysis of the reaction-specifc optimization trajectories (see Supplementary Information \cref{subsubsec:dataset_details,subsubsec:add_results_augmented}) shows that these gains are driven by the borylation and deoxyfluorination reactions, where outcomes depend more strongly on specific condition–substrate pairs. 
These observations indicate the benefit of incorporating lookahead approaches in scenarios with stronger interaction effects, or when fine-tuning conditions near the optimum\citep{toscano-palmerin_bayesian_2022}.

Overall, the results suggest that efficient generality-oriented optimization can be achieved using sequential acquisition strategies.
Given their substantially lower computational cost (\textit{e.g.}, ${\sim}1000 \times$ for \gls{seqonela} \textit{vs.} \gls{jointtwola} on these datasets, see Supplementary Information \cref{subsec:BO_benchmark}), we conclude that sequential strategies provide a favorable trade-off between optimization performance and computational efficiency.
We therefore focus on sequential strategies in the remainder of this work. 

\subsubsection*{Substrate acquisition: Uncertanty-guided selection improves efficiency}

Given that sequential strategies perform comparably to joint approaches, we  next evaluate how substrate selection within this setting affects optimization efficiency. 
Specifically, we compare two perviously used \emph{acquisition functions} for substrate selection: posterior variance-based selection (\gls{pvfunc}) \citep{angello_closed-loop_2022} and random substrate selection (\gls{rafunc}) \citep{wang_identifying_2024}.

\begin{figure*}[t]
    \centering
    \includegraphics[width=\linewidth]{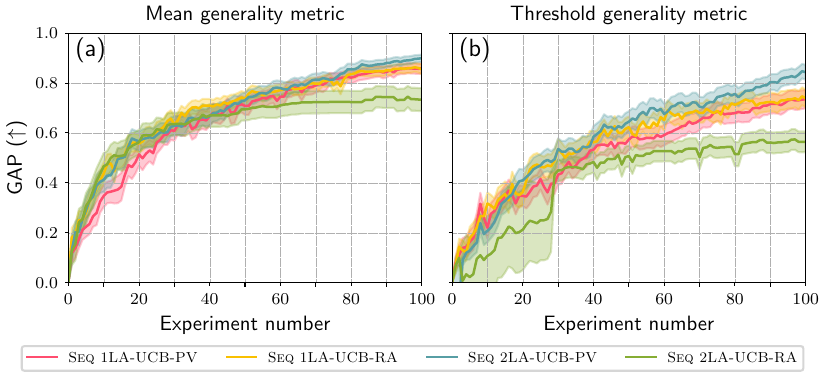}
    \vspace{-1em}
    \caption{
        Optimization trajectories, averaged over all datasets, for the \gls{seqonela} and \gls{seqtwola} acquisition strategies with \gls{pvfunc} and \gls{rafunc} substrate acquisition functions, on the \gls{mean} (a) and \gls{threshold} (b) generality metrics.
    }
    \label{fig:task-selection}
\end{figure*}

Interestingly, we observe different effects of the substrate acquisition function on one- and two-step lookahead strategies (see \cref{fig:task-selection}).
For the one-step lookahead strategy, \gls{pvfunc} and \gls{rafunc} yield similar performance.
In contrast, for the two-step lookahead strategy, \gls{pvfunc} consistently achieves higher \gls{gap} values than \gls{rafunc} for both the \gls{mean} and \gls{threshold} metrics.
These performance difference are most apparent in the later stages of optimization.
Again, reaction-specific analysis reveals that this effect is primarily driven by the deoxyfluorination and borylation reactions.
As discussed above, reaction outcomes in these cases depend more strongly on specific condition–substrate interactions, increasing the importance of informed substrate selection.

Overall, these results indicate that uncertainty-guided substrate selection improves the efficiency of generality-oriented optimization, particularly in combination with two-step lookahead strategies. 
Compared to "myopic" one-step lookahead, these strategies are highly sensitive to the accuracy of predicted future outcomes.
We hypothesize that, in this scenario, reducing uncertainty in key regions of condition–substrate space improves the reliability of these predictions, better resolves condition-substrate interactions, and therefore enhances optimization performance.

\subsubsection*{Condition selection: Explorative acquisition is critical for efficient optimization}

After investigating the role of substrate acquisition, we now examine the condition acquisition function in an analogous manner.
Importantly, we expect that condition selection will have the largest influence on optimization performance, given that sequential strategies that select conditions before substrates achieve performance that is \textit{on par} with joint approaches.

To test this hypothesis, we vary the exploration/exploitation-tradeoff in the \gls{ucbfunc} acquisition function via the $\beta$ parameter, considering values of 0.1 (exploitative), 0.5 (balanced trade-off, as used above), and 5.0 (explorative).

\begin{figure*}[t]
    \centering
    \includegraphics[width=\linewidth]{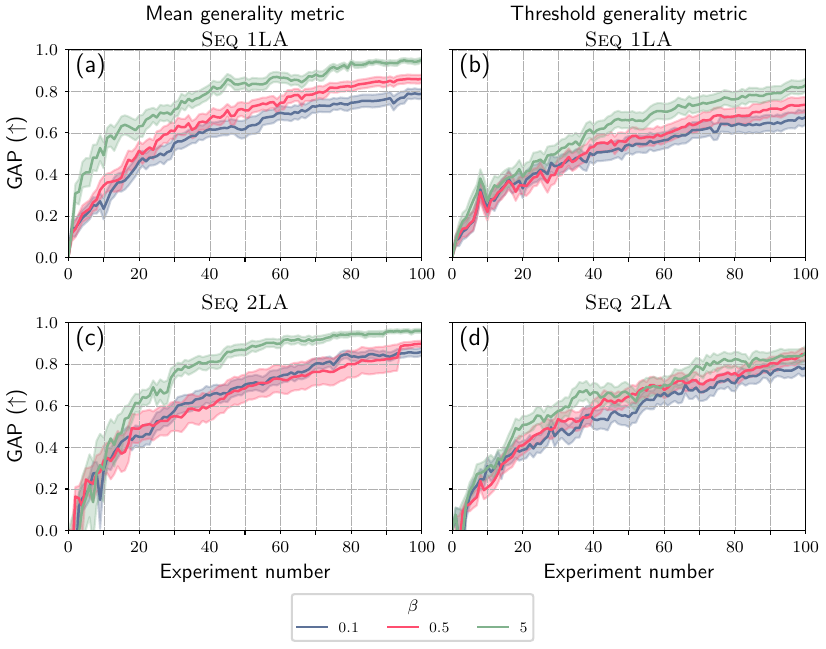}
    \vspace{-1em}
    \caption{
        Optimization trajectories, averaged over all datasets, for the \gls{seqonela} (a,b) and \gls{seqtwola} (c,d) acquisition strategies with varying exploration/exploitation of the conditions acquisition function, on the \gls{mean} (a,c) and \gls{threshold} (b,d) generality metrics.
    }
    \label{fig:parameter-selection}
\end{figure*}

The results show that condition selection is indeed the central driver of optimization performance.
In particular, more explorative settings (\textit{i.e.}, higher values of $\beta$) consistently yield higher \gls{gap} values, for both the one- and two-step lookahead acquisition strategies (\cref{fig:parameter-selection}).
This indicates that broader exploration of condition space is required for the efficient identification of general conditions.

A likely explanation is that exploration improves coverage of the condition space, which is particularly important in the partial monitoring setting. 
Since the generality objective cannot be directly observed, but needs to be inferred from a surrogate model, insufficient exploration can lead to biased or incomplete estimates of the true generality function. 
Increased exploration can improve the quality of the inferred optimum,\citep{lattimore_exploration_2020, kirschner_linear_2023}, resulting in more sample-efficient global optimization.

\subsubsection*{Insights-derived acquisition strategy sets new state-of-the-art}

Building on the insights gained above, we eventually benchmark the best-performing acquisition strategies identified with \gls{currybo} against previously reported approaches in the chemistry domain. 
Specifically, we compare the top-performing strategy (\gls{seqtwolaucbfivepv}) with the methods proposed by Angello \textit{et al.} (corresponding to \gls{seqonelapipv} in the \gls{currybo} framework) \citep{angello_closed-loop_2022} and Wang \textit{et al.} (\gls{bandit}) \citep{wang_identifying_2024}.
As the available implementation of \gls{bandit} does not support the \gls{threshold} metric, we include it only for the \gls{mean} metric.
As baselines, we compare against optimizing on a single model substrate (\gls{seqonelaucbsingle}) \citep{wagen_screening_2022} representing common experimental practice, as well as a fully random baseline (\gls{random}).

\begin{figure*}[t]
    \centering
    \includegraphics[width=\linewidth]{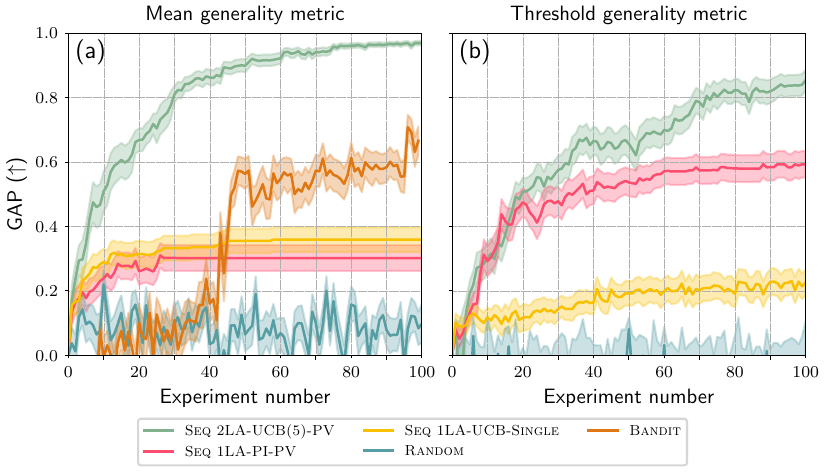}
    \vspace{-1em}
    \caption{
        Optimization trajectories of different algorithms for generality-oriented optimization previously reported in the chemical domain. Note that the available implementation of the \gls{bandit} algorithm is incompatible with the threshold aggregation function.
    }
    \label{fig:chemistry-algos}
\end{figure*}

The results show that the best \gls{currybo}-identified strategy, \gls{seqtwolaucbfivepv}, consistently achieves higher \gls{gap} values than all other algorithms throughout the optimization (\cref{fig:chemistry-algos}).
Comparison against the common practice of optimizing on a single model substrate highlights the importance of multi-substrate optimization. 
While the single-substrate strategy initially increases the \gls{gap} value by optimizing conditions for that very substrate, performance quickly plateaus at suboptimal levels. 
This plateau shows that the identified conditions do not generalize well to other substrates, reinforcing the importance of explicitly optimizing for generality. 

Notably, the \gls{bandit} algorithm from Wang \textit{et al.} \citep{wang_identifying_2024} performs similarly to the \gls{random} baseline during the early stages of optimization.
The reason for this behaviour is that every ``arm'' of the \gls{bandit} (\textit{i.e.}, each possible set of reaction conditions) must be evaluated at least once during an initialization phase before meaningful experiment planning can occur.
During this phase, condition selection is effectively random.
This inherent limitation becomes more pronounced in larger search spaces. While the original study from Wang \textit{et al.} considered problems with 4, 20, and 23 sets of conditions, our benchmarks include on larger, and arguably more realistic search spaces of 20, 43, 46 and 96 possible conditions.
As a result, performance improvements in the \gls{bandit} approach are only observed after approximately 50 experiments, when initialization is complete for most benchmark problems (see \cref{fig:chemistry-algos}a). 
This extended initialization phase leads to a large number of potentially uninformative experiments, limiting the practicality of the bandit approach in larger condition spaces. 
However, once the initialization phase is completed, the method achieves competitive performance. 

Comparing the \gls{bo}-based strategy of Angello \textit{et al.} \citep{angello_closed-loop_2022} with the best-performing \gls{currybo} strategy illustrates the benefit of systematically exploring the design space of acquisition strategies. 
Although both approaches employ sequential condition--substrate selection, the use of two-step lookahead and a more explorative condition acquisition function leads to substantially improved sample efficiency. In contrast, the one-step lookahead strategy with a more exploitative acquisition function (\gls{pifunc}) converges prematurely, resulting in lower \gls{gap} values. 
As mentioned above, this behavior is likely attributed to insufficient exploration of the condition space, leading to convergence to locally optimal but non-general conditions (see Supplementary Information \cref{subsubsec:add_results_augmented} for a detailed discussion).

More broadly, these results underline the value of a flexible framework that enables modular composition and systematic evaluation of different algorithmic strategies. At the same time, they re-emphasize the need for practical benchmark tasks, acknowledging that such benchmark suites remain an imperfect proxy for the diversity of chemical reaction optimization.  
%!TEX root=../main.tex

\section*{Discussion}

We have introduced \gls{currybo}, a modular framework for generality-oriented optimization that enables the independent specification and systematic evaluation of generality metrics, surrogate models, and acquisition strategies.
This modularity allows for a structured exploration of the algorithmic design space, which is typically difficult to achieve in practical application scenarios. 

Using \gls{currybo}, we investigated the impact of key components of the optimizer across multiple datasets from chemical reaction optimization. This analysis revealed a consistent set of design principles for generality-oriented optimization. In particular, we find that (i) optimization over multiple substrates improves the transferability of identified conditions; (ii) sequential acquisition strategies perform comparably to more complex joint approaches, and (iii) efficient optimization is achieved by combining explorative condition selection with uncertainty guided substrate selection. Together, these components define an effective acquisition strategy that sets a new state-of-the-art approach across all benchmarks. 

Importantly, these insights have been enabled by the ability to disentangle and recombine individual components of the optimization pipeline. 
The results therefore highlight the value of a modular formulation of generality-oriented optimization, not only for identifying improved strategies but also for understanding the underlying factors that govern optimization performance.
While the observed improvements depend on the structure of the underlying reaction space, particularly the strength of condition–substrate interactions, the framework provides a general methodology for adapting optimization strategies to specific problem settings. 
In this sense, \gls{currybo} serves both as a practical tool for reaction optimization and as a platform for further methodological development.

Looking forward, the availability of such a framework, together with open benchmark datasets, may enable more systematic investigations into generality-oriented optimization, and facilitate the integration into chemical laboratories.
Beyond chemical reaction optimization, the underlying problem of identifying parameters that generalize across varying conditions arises in many domains, including hyperparameter optimization in machine learning \citep{toscano-palmerin_bayesian_2022} and the tuning of sensing parameters in multi-analyte detection systems \citep{guntner_breath_2019}. 
We therefore expect that the principles and tools introduced here may find broader application in related optimization problems across science and engineering.
%!TEX root=../main.tex

\section*{Methods}

\subsection*{Generality-oriented Optimization} \label{subsec:Generality-BO}

\subsubsection*{Extended Problem Formulation} \label{subsubsec:algo_outline}

As outlined above, we consider a black-box function $f: \gX \times \gW \rightarrow \mathbb{R}$ in joint space $\gX \times \gW$. %, where the outcome is a real number (\textit{e.g.} reaction yield), $\rvx \in \gX$ can be continuous, discrete or mixed-variable and $\gW = \{ \rvw_i \}_{i=1}^n$ is a discrete task space of size $n$.
To underline the utility of \gls{currybo} beyond reaction optimization, we refer to the conditons ($\rvx$) as parameters, and the substrates ($\rvw$) as tasks.
Each evaluation of $f$ is expensive and does not provide gradient information.
Let \(\mathrm{curry}\) be a currying operator on the second argument, defined by $\mathrm{curry}(f): \gW \rightarrow \big( \gX \rightarrow \mathbb{R} \big)$.
Then, for some \emph{specific} $\rvw_{\mathrm{i}} \in \gW$, evaluating \(\mathrm{curry}(f)(\rvw_{\mathrm{i}})\) yields a new, partially-applied function $f_{\rvw_{\mathrm{i}}}: \gX \rightarrow \mathbb{R}$ %$f (\,\boldsymbol{\cdot}\,; \rvw): \gX \rightarrow \mathbb{R}$
, where $f_{\rvw_{\mathrm{i}}}(\rvx) = f(\rvx, \rvw_{\mathrm{i}})$. 
This is motivated by the fact that each member of this family of functions $f_{\rvw}: \gX \rightarrow \mathbb{R}$ corresponds to a function that can be evaluated experimentally. 
%In other words, all observable functions can be described through an $n$-sized set $\gF = \{ f_{\rvw_i}: \gX \rightarrow \mathbb{R} \}_{i=1}^{n}$.
%In the context of reaction condition optimization $\gF$ consists of all functions that describe the reaction outcome for each substrate.
In the reaction context, evaluation of a specific $f_{\rvw_i}(\rvx)$ then corresponds to measuring the reaction outcome of a substrate (described by $\rvw_i$) under specific reaction conditions $\rvx$.

In generality-oriented optimization, the goal is to identify the optimum $\hat{\rvx} \in \gX$ that is generally optimal across $\gW$, meaning $\hat{\rvx}$ maximizes a user-defined generality metric $\phi$ over all $\rvw \in \gW$. 
\begin{equation} \label{eq:aggrate-BO}
\hat{\rvx} = \underset{\rvx \in \gX}{\text{argmax}} \ \phi(\rvx) :=  \underset{\rvx \in \gX}{\text{argmax}} \ \phi\big(f_{\rvw}(\rvx), \gW\big) 
\end{equation}
While \eqref{eq:aggrate-BO} appears like a standard global optimization problem over \(\gX\), evaluating \(\phi(\rvx)\) itself is intractable due to the generality metric over \(\gW\).
To evaluate \(\phi(\rvx)\) on a single \(\rvx\), one must perform \(n\)-many expensive function evaluations to first obtain \(\{ f_{\rvw_i}(\rvx) \}_{i=1}^n\).
%Ideally, the number of such function evaluations should be minimized.
%Thus, this setting differs from the conventional global optimization problem, due to its \emph{partial observation} nature:
%One can only estimate \(\phi(\rvx)\) via a subset of observations \(\{ f_{\rvw_j}(\rvx)\}_{j=1}^m\) where \(m < n\).
To maximize sample efficiency, an optimizer should always recommend a new pair $(\rvx_{k+1}, \rvw_{k+1})$ to evaluate next---in other words: $\phi(\rvx_{k+1})$ is only observed partially via a single evaluation of \(f\).
Treating this in the conventional framework of \gls{bo}, we can build a probabilistic surrogate model $g(\rvx, \rvw)$ from all $k$ available observations $\gD = \{(\rvx_i, \rvw_i, f_{\rvw_i}(\rvx_i)\}_{i=1}^k$, referred to as $p\big(g_k (\rvx, \rvw) \mid \mathcal{D}\big)$. 
From the posterior distribution over $g$, a posterior distribution over $\phi$ can be estimated for any functional form of $\phi$ via Monte-Carlo integration \citep{balandat_botorch_2020} (see Supplementary Information \cref{App:acqf} for further details). 

Unlike the conventional \gls{bo} case, we now need a specific acquisition policy $A$ to decide at which $\rvx \in \gX$ \emph{and} $\rvw \in \gW$, $(\rvx_{k+1},\rvw_{k+1})$, the objective function \(\phi(\rvx)\) should be partially evaluated.
%\(A\) plays an important role since it must respect the partial observability constraint, \emph{i.e.} propose \emph{a single} $(\rvx_{k+1},\rvw_{k+1})$ at each step such that the general (over \emph{all} \(\rvw_i\)'s) optimum \(\hat{\rvx}\) can be reached in as few steps as possible.
Given the pair $(\rvx_{k+1},\rvw_{k+1})$, the objective \(\phi(\rvx_{k+1})\) is partially observed, $\gD$ is updated, and the steps are repeated until the budget is exhausted.
Owing to the partial monitoring scenario \citep{rustichini_minimizing_1999, lattimore_cleaning_2019, lattimore_bandit_2020}, the final optimum after a budget of $K$ experiments, $\hat{\rvx}_K$, is returned as the $\rvx \in \gX$ that maximizes the mean of the predictive posterior of $\phi: p\big(\phi (\rvx) \mid \mathcal{D}_K\big)$.
These steps are summarized in \cref{alg:genbo_general}.

\begin{algorithm}[t]

\caption{Generality-oriented Bayesian optimization} \label{alg:genbo_general}
\footnotesize
\begin{algorithmic}[1]
\Require
\Statex Set of observable functions  $\{ f_{\rvw_i}(\rvx): \gX \rightarrow \mathbb{R} \}_{i=1}^{n}$
\Statex Initial dataset $\mathcal{D}_k = \big\{ \rvx_j, \rvw_j, f_{\rvw_j}(\rvx_j)  \big\}_{j=1}^k$
\Statex Generality metric $\phi\big(f_{\rvw}(\rvx), \gW\big)$
\Statex Surrogate model $g(\rvx, \rvw)$ and acquisition policy $A$
\Statex Budget $K$

\vspace{0.2cm}

\While {\(k \le K \)}
\State Compute posterior $p\big(g_k (\rvx, \rvw) \mid \mathcal{D}_k\big)$
\State Acquire $\rvx_{k+1}, \rvw_{k+1} = A\Big(p\big(g_k (\rvx, \rvw) \mid \mathcal{D}_k\big), \phi\big(f_{\rvw}(\rvx), \gW\big) \Big)$
\State Observe $f_{\rvw_{k+1}}(\rvx_{k+1})$
\State Update $\mathcal{D}_{k+1} = \mathcal{D}_k \bigcup \big\{ (\mathbf{x_{k+1}}, \rvw_{k+1}, f_{\rvw_{k+1}}(\rvx_{k+1}))  \big\}$
\State $k = k+1$
\EndWhile

\vspace{0.2cm}

\State \Return $\hat{\rvx}_K = \underset{\rvx \in \gX}{\mathrm{argmax}} \ \mathbb{E} \Big[ p\big(\phi (\rvx) \mid \mathcal{D}_K\big) \mid \rvx \Big]$
\end{algorithmic}
\end{algorithm}

\subsubsection*{Acquisition Strategies for $\rvx_{k+1}$ and $\rvw_{k+1}$}\label{subsubsec:acquisition_strategies_main}
As outlined above, the efficiency of generality-oriented optimization depends on the selection of $\rvx_{k+1}$ and $\rvw_{k+1}$. 
Given a posterior distribution $p\big(g_k (\rvx, \rvw) \mid \mathcal{D}\big)$ and a generality metric $\phi\big(f_{\rvw}(\rvx), \gW\big)$, any acquisition policy should determine $\rvx_{k+1}$ and $\rvw_{k+1}$, which formally requires optimization over $\gX \times \gW$.
Assuming weak coupling between $\gX$ and $\gW$, we can formulate a sequential acquisition policy (as outlined in \cref{alg:sequential_acquisition}). 
First, $\rvx_{k+1}$ is acquired by optimizing an $\rvx$-specific acquisition function $\alpha_x$ over $\phi: p\big(\phi (\rvx) \mid \mathcal{D}_k\big)$. 
Second, a $\rvw$-specific acquisition $\alpha_w$ is optimized over this posterior distribution at $\rvx_{k+1}$. Notably, in this setting, established acquisition functions can be used for both $\alpha_x$ and $\alpha_w$.
For those acquisition functions that require the current maximum as input (\textit{e.g.} \gls{pifunc} or \gls{eifunc}), this maximum is taken as the $\rvw$ that maximizes the posterior mean over $\phi: p\big(\phi (\rvx) \mid \mathcal{D}_k\big)$.

However, the decoupling of $\gX$ and $\gW$ is a strong assumption. Thus, we also evaluate algorithms that identify $\rvx_{k+1}$ and $\rvw_{k+1}$ through joint optimization over $\gX \times \gW$ (\cref{alg:joint_acquisition}), requiring a two-step lookahead acquisition function $\alpha'$:
\begin{equation} \label{eq:two-step-la}
\footnotesize
\alpha' (\rvx_{k+1}, \rvw_{k+1}) = \alpha \Bigg[ \underset{\rvx_{k+2} \in \gX}{\mathrm{argmax}} \ \alpha_{\mathrm{final}} \bigg(p\Big(\phi\big(\rvx_{k+2} \big) \mid \mathcal{D}^*_{k+1} \Big)\bigg)\Bigg]
\end{equation}
where $\alpha$ is a classical one-step lookahead acquisition function, which is evaluated at $\rvx_{k+2} \in \gX$ which maximizes the acquisition function for making the final decision $\alpha_{\mathrm{final}}$ (in our case: greedy acquisition) over a fantasy posterior distribution $p\Big(\phi\big(\rvx\big) \mid \mathcal{D}^*_{k+1} \Big)$. This distribution is obtained by conditioning the existing posterior on a new fantasy observation at ($\rvx_{k+1}, \rvw_{k+1}$), which is sampled from the predicted distribution of the posterior $p\big(g_k (\rvx, \rvw) \mid \mathcal{D}_k\big)$.
An implementation of \cref{eq:two-step-la} using Monte-Carlo integration is given in \cref{alg:joint_acquisition}.
The next values $\rvx_{k+1}$ and $\rvw_{k+1}$ are then acquired by optimizing $\alpha'$ in the joint input space $\gX \times \gW$.

\subsection*{Chemical Reaction Optimization Datasets}

In our experiments, we consider four chemical reaction datasets stemming from high-throughput experimentation: (1) Pd-catalyzed carbon-heterocoupling \citep{buitrago_santanilla_nanomole-scale_2015}, (2) \textit{N},\textit{S}-acetal formation \citep{zahrt_prediction_2019}, (3) borylation \citep{stevens_advancing_2022}, and (4) deoxyfluorination \citep{nielsen_deoxyfluorination_2018}. 
Each dataset evaluates the optimization of a chemically relevant reaction outcome (such as enantioselectivity given as $\Delta\Delta G^{\ddagger}$, yield, or starting material conversion), and contains an experimental dataset of substrates, conditions and measured outcomes.
The number of condition combinations range from 20 to 96, with the number of substrates ranging from 31 to 75.
Extensive analysis of the benchmark problems can be found in the Supplementary Information \cref{subsubsec:dataset_details}.
At this stage, it should be noted that, while widely used as such, the problems have not been designed as benchmarks for reaction condition optimization. 
To mitigate the well-known bias of HTE datasets towards high-outcome experiments \citep{strieth-kalthoff_machine_2022, beker_machine_2022}, we additionally augment the search space to incorporate larger domains of low-outcome results using a chemically sensible space expansion workflow (see the Supplementary Information \cref{subsubsec: augmentation} for details).

\subsection*{Grid Search for Upper-Bound Analysis}

To analyze the utility of considering multiple substrates in an optimization campaign, we performed exhaustive grid search on the described datasets. 
For each, the substrates were split into an initial train and test set among the substrates.
In total, 30 different train/test splits were generated.
The obtained train set was further subsampled into smaller training sets with varying sizes.
Sampling among the substrates in the train set was performed either randomly (main text), or by farthest point sampling or Tanimoto similarity-based sampling (see additional results in Supplementary Information \cref{subsubsec:add_results_data-analysis}).
For each subsampled training set, the most general conditions were identified via exhaustive grid search.
The general reaction outcome, as specified by the generality metric, is evaluated for these conditions on the held-out test set.
Further, this general reaction outcome was min-max scaled from 0 (worst) to 1 (best) to give a dataset-independent generality score, \textit{i.e.}, $\mathrm{generality\, score} = \frac{\phi^{\mathrm{test}}(\rvx_{\mathrm{search}}) - \phi^{\mathrm{test}}(\rvx_{\mathrm{min}})}{\phi^{\mathrm{test}}(\rvx_{\mathrm{max}}) - \phi^{\mathrm{test}}(\rvx_{\mathrm{min}})}$, where $\phi^{\mathrm{test}}$ is the generality metric applied on the substrates of the test set.
$\rvx_{\mathrm{search}}$, $\rvx_{\mathrm{min}}$, and $\rvx_{\mathrm{max}}$ are the conditions with the highest generality on the train substrates, the conditions with the lowest generality on the test substrates and the conditions with the highest generality on the test substrates, respectively.

\subsection*{Optimization Algorithms} \label{subsec:strategies}
Using the datasets outlined above, we perform systematic evaluations of multiple policies for identifying general optima.
Each policy is evaluated under two different generality metrics: the \gls{mean} and the \gls{threshold} generality metrics described above (see Supplementary Information \cref{subsubsec:policies} for further details and \cref{subsubsec: Aggregation_functions} for additional practically relevant generality metrics).
Molecules are represented using Morgan Fingerprints \citep{morgan_generation_1965} with 1024 bits and a radius of 2, generated using \texttt{RDKit} \citep{landrum_rdkit_2023}.
In all \gls{bo} experiments, we use a \gls{gp} surrogate, as provided in \textit{BoTorch} \citep{balandat_botorch_2020}, with the Tanimoto kernel from \textit{Gauche} \citep{griffiths_gauche_2023}.
It should be noted that this choice slightly differs from the implementation by Angello \textit{et al.} \citep{angello_closed-loop_2022}, who used a Mátern32 kernel on Morgan fingerprints.
While Kernel choice can lead to slight variations in the results, our goal is to compare the experimental decision making algorithms on a comparable setting.
Therefore, we used the Tanimoto Kernel, which is specifically designed for molecular fingerprints, across all experiments. 
For each reaction, we provide the mean and standard error of the \gls{gap} value \citep{jiang_binoculars_2020} over 30 independent runs, each with different substrates and initial conditions.

\subsubsection*{Details on Bandit Algorithm Benchmarking}

The \gls{bandit} algorithm \citep{wang_identifying_2024} was benchmarked using the \texttt{UCB1Tuned} version, as implemented by the authors. 
For each reaction, we selected the substrates and initial conditions across 30 independent runs consistent to all other optimization algorithms, and report the same metrics. 
To select the optimum $\hat{\rvx}$ value at each step $k$, we relied on the authors' definition of the best arm as the most sampled arm at step $k$\citep{wang_identifying_2024}.

\section*{Acknowledgements}

This publication was created as part of NCCR Catalysis (grant numbers 180544 and 225147), a National Centre of Competence in Research funded by the Swiss National Science Foundation. S.P.S. thanks Aline Hartgers for fruitful discussions.
E.M.R. thanks the Vector Institute. S.X.L. acknowledges support from Nanyang Technological University, Singapore and the Ministry of Education, Singapore for the Overseas Postdoctoral Fellowship. A.A.-G. thanks Anders G. Fr{\o}seth for his generous support. A.A.-G. also acknowledges the generous support of Natural Resources Canada and the Canada 150 Research Chairs program. This research is part of the University of Toronto’s Acceleration Consortium, which receives funding from the Canada First Research Excellence Fund (CFREF). Resources used in preparing this research were provided, in part, by the Province of Ontario, the Government of Canada through CIFAR, and companies sponsoring the Vector Institute.

\section*{Code and Data Availability}

All datasets (augmented and non-augmented), code for the application of \textit{CurryBO} and reproduction of the results are available under: \href{https://github.com/digital-chemistry-laboratory/currybo}{https://github.com/digital-chemistry-laboratory/currybo}.

\bibliographystyle{unsrtnat}
\bibliography{references_cleaned}

\newpage
\setcounter{page}{1}
\renewcommand{\thepage}{S\arabic{page}}
%!TEX root=../main.tex

\appendix
\onecolumn
\setcounter{section}{1}

\renewcommand{\thesection}{\arabic{section}}
\renewcommand{\thesubsection}{\thesection.\arabic{subsection}}
\renewcommand{\thesubsubsection}{\thesubsection.\arabic{subsubsection}}

\renewcommand{\thefigure}{S\arabic{figure}}
\setcounter{figure}{0}
\renewcommand{\thetable}{S\arabic{table}}
\setcounter{table}{0}
\renewcommand{\thealgorithm}{S\arabic{algorithm}}
\setcounter{algorithm}{0}
\section*{Supporting Information}

\subsubsection*{Global Optimization}

Global black-box optimization is concerned with finding the optimum of an unknown objective function $f(\rvx)$: 
\begin{equation}
\hat{\rvx} = \underset{\rvx \in \gX}{\text{argmax}} \ f(\rvx)
\end{equation}
Suppose $f(\rvx)$ is a function that (a) is not analytically tractable, (b) is very expensive to evaluate, and (c) can only be evaluated without obtaining gradient information. In this scenario, \gls{bo} has emerged as a ubiquitous approach for finding the global optimum $\hat{\rvx} \in \gX$ in a sample-efficient manner \citep{garnett_roman_bayesian_2023}. 
The working principle of \gls{bo} involves a probabilistic surrogate model $g(\rvx)$ to approximate $f(\rvx)$, which can be used to compute a posterior predictive distribution over $g$ under all previous observations $\gD = \{(\rvx_i, f(\rvx_i)\}_{i=1}^k$. 
The most prominent choice for $p(g(\rvx) \mid \gD)$ are \gls{gp}s \citep{rasmussen_gaussian_2006}, with various types of Bayesian neural networks becoming increasingly popular in the past decade \citep{hernandez-lobato_parallel_2017, kristiadi_promises_2023, li_study_2024, kristiadi_sober_2024}. 
Based on the predictive posterior, an acquisition function $\alpha$ over the input space $\gX$ is used to decide at which $\rvx_{k+1} \in \gX$ the objective function should be evaluated next. 
Key to the success of \gls{bo} is the implicit exploitation–exploration tradeoff in $\alpha$, which makes use of the posterior distribution $p(g(x) \mid \gD)$ \citep{mockus_bayesian_1975}.
Common choices of $\alpha$ are \gls{ucb} \citep{kaelbling_associative_1994, kaelbling_associative_1994-1, agrawal_sample_1995}, \gls{ei} \citep{jones_efficient_1998}, \gls{kg} \citep{gupta_bayesian_1994, frazier_knowledge-gradient_2008, frazier_knowledge-gradient_2009} or \gls{ts} \citep{thompson_likelihood_1933}.
The hereby selected $\rvx_{k+1}$ is evaluated experimentally, resulting in \(f(\rvx_{k+1})\), and the described procedure is repeated until a satisfactory outcome is observed, or a budget is exhausted.

\subsection{Bayesian Optimization under partial monitoring} \label{subsec:Generality-BO}

\subsubsection{Generality metrics} \label{subsubsec: Aggregation_functions}

The generality metric $\phi$ is a user-defined property that determines how the objective is calculated across all tasks (substrates). 
Through the choice of the generality metric, prior knowledge and preferences about the specific optimization problem at hand can be included. 
In the optimization literature, if $\phi$ contains a sum over all $f_{\rvw_i}(\rvx)$ with $\rvw_i \in \gW$, it can be referred to as the optimization of integrated response functions \citep{williams_sequential_2000}, optimizing an average over multiple tasks \citep{swersky_multi-task_2013}, or optimization with expensive integrands \citep{toscano-palmerin_bayesian_2022}.
\citet{toscano-palmerin_bayesian_2022} propose a \gls{bo} approach, including a joint acquisition over $\gX \times \gW$ with the goal of maximizing the value of information. 
In the \gls{currybo} framework, this acquisition strategy corresponds to \gls{jointtwolaei}.

Another generality metric that was considered in literature is $\phi$ corresponding to the $min$ operation (see Minimum generality metric below) has been discussed as distributionally robust \gls{bo} \citep{bogunovic_adversarially_2018, kirschner_distributionally_2020, nguyen_distributionally_2020, husain_distributionally_2023}.
In this work, the following generality metrics are evaluated:

\textbf{\gls{mean} generality metric}

\begin{equation}
\phi(f_{\rvw}(\rvx), \mathcal{W}) = \frac{1}{|\mathcal{W}|} \sum_{\rvw \in \mathcal{W}} f_{\rvw}(\rvx) = \frac{1}{n} \sum_{i=1}^{n}f_{\rvw_i}(\rvx)
\end{equation}

\textbf{\gls{threshold} generality metric}

\begin{equation}
\phi\big(f_{\rvw}(\rvx\big), \mathcal{W}) = \sum_{\rvw \in \mathcal{W}} \sigma\big(f_{\rvw}(\rvx) - f_{\mathrm{thr}}\big) = \sum_{i=1}^{n} \sigma\big(f_{\rvw_i}(\rvx) - f_{\mathrm{thr}}\big),
\end{equation}

where $\sigma$ denotes the sigmoid function $\sigma(x) = \frac{1}{1 + e^{- k \cdot x}}$, where we choose the hyperparameter $k=300$ for a steep threshold function. As described above, other generality metrics also have practical use-cases, for example:

\textbf{Mean Squared Error (MSE) generality metric}

\begin{equation}
\phi\big(f_{\rvw}(\rvx), \mathcal{W}\big)  = - \frac{1}{|\mathcal{W}|} \sum_{\rvw \in \mathcal{W}} \big(f^{\mathrm{opt}}_{\rvw}(\rvx) - f_{\rvw}(\rvx)\big)^2 = - \frac{1}{n} \sum_{i=1}^n \big(f^{\mathrm{opt},i} - f_{\rvw_i}(\rvx)\big)^2
\end{equation}

\textbf{Minimum generality metric}

\begin{equation}
\phi\big(f_{\rvw}(\rvx), \mathcal{W}\big)  = \min_{\rvw_i \in \mathcal{W}} f_{\rvw_i}(\rvx)
\end{equation}

The above definitions assume that all $f_{\rvw_i}(\rvx)$ have the same range, and that the optimization problem is formulated as a maximization problem.
The \gls{mean} and \gls{threshold} metrics have been chosen for our experiments, as these metrics have been discussed in the chemical literature on general reaction conditions \citep{angello_closed-loop_2022, betinol_data-driven_2023, wang_identifying_2024, gallarati_genetic_2024}

\subsubsection{Acquisition Strategies and the Sample Average Approximation} \label{App:acqf}

For the evaluation of posterior distributions, and the calculation of acquisition function values, we use the \gls{saa}, as introduced by \citet{balandat_botorch_2020}. 
From a posterior distribution at time point $k$, $p\big(g_k(\rvx)\big)$, $M$ posterior samples $\zeta_{m}(\rvx) \sim p\big(g_k(\rvx)\big)$ are drawn. 
These posterior samples can be used to estimate the posterior distribution, and to calculate acquisition function values as expectation values $\mathbb{E}_M$ over all $M$ samples.

Herein, we use the following common acquisition functions: 
\begin{itemize}
    \item Upper Confidence Bound: $\text{\gls{ucbfunc}}(\rvx) = \mathbb{E}_M\big(\zeta_{m}(\rvx)\big) + \beta \cdot \mathbb{E}_M\big(\zeta_{m}(\rvx) - \mathbb{E}_M(\zeta_{m}(\rvx))\big)$.
    \item Probability of Improvement: $\text{\gls{pifunc}}(\rvx) = \mathbb{P}_M\big(\zeta_{m}(\rvx) > f^*\big)$, where $f^{*}$ is the best value observed so far.
    \item Posterior Variance: $\text{\gls{pvfunc}}(\rvx) = \mathbb{E}_M\big(\zeta_{m}(\rvx) - \mathbb{E}_M(\zeta_{m}(\rvx))\big)$. 
    \item Random Selection \gls{rafunc}, where the acquisition function value is a random number. 
\end{itemize}

Moreover, we evaluate the optimization performance using an implementation of two-step lookahead acquisition function $\alpha^{*}$. 
The acquisition function value of $\alpha^{*}$ at a location $\mathbf{x_0}$ is estimated as follows (see also \cref{alg:two_step}): For each of the $M$ posterior samples $\zeta_m(\mathbf{x_0}) \sim p\big(g_k(\mathbf{x_0})\big)$, a fantasy posterior distribution $p'\big(\phi(g_{k+1}(\mathbf{x_0}))\big)$ is generated by conditioning the posterior on the new observation $(\mathbf{x_0}, \zeta_M(\mathbf{x_0}))$ and generality metric.
From this fantasy posterior distribution, the values of the inner acquisition function $\alpha_m$ can be computed and optimized over $\rvx \in \mathcal{X}$. 
The final value of the two-step lookahead acquisition function is returned as $\alpha^{*}(\rvx_0) = \frac{1}{M} \sum_{m=1}^M \alpha_m$.

\begin{algorithm} 

\caption{
Two-step lookahead acquisition function using the sample average approximation.}\label{alg:two_step}

\begin{algorithmic}[1]
\Require
\Statex input space $\mathcal{X}$
\Statex location $\mathbf{x_0}$ at which to evaluate the two-step lookahead acquisition function
\Statex generality metric $\phi\big(f(\rvx; \rvw), \mathcal{W}\big)$
\Statex posterior distribution  $p\big(g_k (\rvx) \mid \mathcal{D}\big)$
\Statex one-step lookahead acquisition function $\alpha(\rvx)$

\vspace{0.2cm}

\State draw $M$ posterior samples $\zeta_m(\mathbf{x_0}) \sim p\big(g_k(\mathbf{x_0})\big)$
\State empty set of fantasy acquisition function values $\mathcal{A} = \{\}$

\For {\(m = 1, \dots, M\)}

 \State compute fantasy posterior $p'(\rvx) = p\Big(\phi\big(g_{k+1} (\rvx) \mid (\mathcal{D} \cup (\mathbf{x_0}, \zeta_m(\mathbf{x_0}))\big)\Big)$
 \State optimize one-step-lookahead acquisition function $\alpha_m = \underset{\rvx \in \mathcal{X}}{\text{max}} \ \alpha(p'(\rvx))$
 \State update $\mathcal{A} = \mathcal{A} \cup \{\alpha_m$\}

\EndFor
\vspace{0.2cm}
\State \Return $\alpha^{*}(\rvx_0) = \frac{1}{M} \sum_{m=1}^M \alpha_m$

\end{algorithmic}
\end{algorithm}

These acquisition functions can be used both in sequential and joint acquisition of the next parameter/task combination, as outlined in \cref{alg:sequential_acquisition,alg:joint_acquisition}, respectively.

\begin{algorithm} 

\caption{Sequential Acquisition Strategy}\label{alg:sequential_acquisition}
\footnotesize
\begin{algorithmic}[1]
\Require
\Statex Posterior distribution  $p\big(g_k (\rvx, \rvw) \mid \mathcal{D}\big)$
\Statex Generality metric $\phi\big(f_{\rvw}(\rvx), \gW\big)$
\Statex Acquisition function $\alpha_x$
\Statex Acquisition function $\alpha_w$

\vspace{0.2cm}

\State Compute posterior $p\big(\phi (\rvx) \mid \mathcal{D}\big) = p \Big(\phi \big(g_k (\rvx, \rvw), \gW\big) \mid \mathcal{D} \Big)$
\State Acquire $\rvx_{k+1} = \underset{\rvx \in \gX}{\text{argmax}} \ \alpha_x\Big(p\big(\phi (\rvx) \mid \mathcal{D}\big)\Big)$
\State Acquire $\rvw_{k+1} = \underset{\rvw \in \gW}{\text{argmax}} \ \alpha_w\Big(p \big(g_k (\rvx_k+1, \rvw) \mid \mathcal{D}\big)\Big)$
\vspace{0.2cm}

\State \Return $\rvx_{k+1}, \rvw_{k+1}$ 

\end{algorithmic}
\end{algorithm}

\begin{algorithm}
\caption{Joint Acquisition Strategy}\label{alg:joint_acquisition}
\footnotesize
\begin{algorithmic}[1]
    \Require
    \Statex Posterior distribution  $p\big(g_k (\rvx, \rvw) \mid \mathcal{D}\big)$
    \Statex Generality metric $\phi\big(f_{\rvw}(\rvx), \gW\big)$
    \Statex Two-step lookahead acquisition function $\alpha'$
    
    \vspace{0.2cm}
    
    \State Compute posterior $p\big(\phi (\rvx) \mid \mathcal{D}\big) = p \Big(\phi \big(g_k (\rvx, \rvw), \gW\big) \mid \mathcal{D} \Big)$
    %\STATE acquire $\rvx_{k+1}, \rvw_{k+1}  = \underset{\rvx, \rvw \in \gX \times \gW}{\text{argmax}} \ \alpha'\Big(p\big(\phi (\rvx) \mid \mathcal{D}\big)\Big)$
    \State Acquire $\rvx_{k+1}, \rvw_{k+1}  = \underset{\rvx, \rvw \in \gX \times \gW}{\text{argmax}} \ \alpha'\Big(\rvx,\rvw\Big)$
    \vspace{0.2cm}
    
    \State \Return $\rvx_{k+1}, \rvw_{k+1}$

\end{algorithmic}
\end{algorithm}

\subsubsection{Benchmarked Optimization Strategies for Selecting $\rvx_{k+1}$ and $\rvw_{k+1}$} \label{subsubsec:policies}

Herein, we outline the use of the investigated acquisition strategies for \gls{bo} under partial monitoring. 
The discussed optimization strategies describe different variations of how to pick the next experiments $\rvx_{k+1}$ and $\rvw_{k+1}$. 

Following the \gls{saa} \citep{balandat_botorch_2020} outlined above, we estimate the predictive posterior distribution $p\big(\phi (\rvx) \mid \mathcal{D}\big)$ as follows: 
For each $\rvw_i \in \mathcal{W}$, $M$ (typically $M = 512$ for one-step lookahead strategies and $M = 3$ for two-step lookahead strategies to reduce computational costs) samples $\zeta_{im}(\rvx) \sim p\big(g_k(\rvx, \rvw_i)\big)$ are drawn from the posterior distribution of the surrogate model.
Applying the generality metric over all $\rvw_i$ yields $M$ samples $\zeta_m (\rvx) \sim p\big(\phi (\rvx) \mid \mathcal{D}\big)$ from the posterior distribution over $\phi(\rvx)$, which can be used for calculating the acquisition function values using the sample-based acquisition function logic, as described in \cref{App:acqf} and in \cref{alg:two_step} for the two-step lookahead acquisition strategy.
With this, we implement and evaluate the acquisition policies in \cref{tab:acquisition_strategies_all}.

\begin{table}[t]
    \footnotesize
    \centering
    \caption{
    Nomenclature and description of the all benchmarked acquisition strategies and acquisition functions, as discussed in the main text and the Appendix. Each experiment is named according to the acquisition strategy used, followed by specifications of the used acquisition functions $\alpha_x$ and $\alpha_w$ or $\alpha$ for sequential and joint acquisitions, respectively.
    As an example, a sequential two-step lookahead acquisition strategy with an Upper Confidence Bound as $\alpha_x$ and Posterior Variance as $\alpha_w$, is referred to as \textsc{Seq 2LA-UCB-PV}.
    }
    \renewcommand{\arraystretch}{1.3}

    \begin{tabularx}{\textwidth}{p{0.55 \textwidth}|X}\toprule
        Acquisition Strategy & Acquisition Function \\ \midrule
        \gls{seqonela}: Sequential acquisition of $\rvx_{k+1}$ and  $\rvw_{k+1}$, each using a one-step lookahead acquisition function. The final $\hat{\rvx}$ is selected greedily. & \gls{ucbfunc}: Upper confidence bound ($\beta = 0.5$). \\
        \gls{seqtwola}: Sequential acquisition of $\rvx_{k+1}$ and  $\rvw_{k+1}$, each using a two-step lookahead acquisition function. The final $\hat{\rvx}$ is selected greedily. & \gls{ucbfuncs}: Upper confidence bound ($\beta = \rvs$). \\
        \gls{jointtwola}: Joint acquisition of $\rvx_{k+1}$ and  $\rvw_{k+1}$ using a two-step lookahead acquisition function. The final $\hat{\rvx}$ is selected greedily. & \gls{pifunc}: Probability of Improvement. \\
        \gls{bandit}: Multi-armed bandit algorithm as implemented by \citet{wang_identifying_2024}. & \gls{pvfunc}: Posterior Variance. \\
        \gls{random}: Random selection of $\hat{\rvx}$. & \gls{rafunc}: Random acquisition.\\
         & \gls{singlefunc}: Selection of the same substrate ($\rvw$) for every iteration. \\
%          & \textsc{Complete}: Selection of every substrate (i.e. every $\rvw \in \gW$) for a selected $\rvx_k+1$. \\
        \bottomrule
    \end{tabularx}
    \label{tab:acquisition_strategies_all}
\end{table}

%The sequential acquisition is described in \cref{alg:sequential_acquisition} and refers to a strategy in which $\rvx_{k+1}$ and $\rvw_{k+1}$ are selected sequentially. In the first step, $\rvx_{k+1}$ is selected by optimizing an $\rvx$-specific acquisition function $\alpha_x$ over $\rvx \in \mathcal{X}$. With the selected $\rvx_{k+1}$ in hand, $\rvw_{k+1}$ is then selected by optimizing an independent, $\rvw$-specific acquisition function over $\rvw \in \mathcal{W}$. With $\alpha_x = \text{PI}$ (Probability of Improvement) and $\alpha_w = \text{PV}$, this would correspond to the strategy described in \citep{angello_closed-loop_2022}. 
%In contrast, the joint acquisition%, as outlined in \cref{alg:joint_acquisition}, 
%refers to a strategy in which $\rvx_{k+1}$ and $\rvw_{k+1}$ are selected jointly through optimization of a two-step lookahead acquisition function (see \cref{alg:two_step} and \cref{App:acqf}).

\subsubsection*{Distinction from Existing Variants of the \gls{bo} Formalism} \label{subsubsec:differentiation}

Despite seeming similarities with \textit{multiobjective}, \textit{multifidelity}, and \textit{mixed-variable} optimization, the generality-oriented approach describes a distinctly different scenario:
\begin{itemize}
    \itemsep 0.25em
    \item{In contrast to \textit{multiobjective} optimization, here, we consider a single optimization objective, \textit{i.e.}, $\phi(\rvx)$. However, this objective can only be observed partially. Whereas the overall optimization problem aims to identify $\hat{\rvx} \in \gX$, finding the next recommended observation requires a joint optimization over $\gX$ and $\gW$.}
    \item{In contrast to \textit{multifidelity} \gls{bo}, the functions parameterized by $\rvw \in \gW$ do not correspond to the same objective with different fidelities. Rather, they are independent functions which all contribute equally to the objective function $\phi(\rvx)$.}
    \item Unlike \textit{mixed-variable} \gls{bo} \citep{daxberger_mixed-variable_2020}, the goal of generality-oriented \gls{bo} is not to find $(\rvx, \rvw)$ that maximizes the objective in the \emph{joint} space. 
    Rather, the goal is to find the set optimum $\hat{\rvx}$ that maximizes $\phi\big(f(\rvx; \rvw), \gW\big)$ over $f(\rvx; \rvw)$. 
    In the case of $\phi$ being a sum, this bears resemblance to maximizing the \emph{marginal} over $\rvx$. 
\end{itemize}

\newpage
\subsection{Extended details on reaction condition dataset} \label{subsec:problem_details}

Four chemical reaction datasets have been considered in this work: (1) reactant conversion optimization for Pd-catalyzed C–heterocoupling \citep{buitrago_santanilla_nanomole-scale_2015}, (2) enantioselectivity optimization for a \textit{N},\textit{S}-acetal formation \citep{zahrt_prediction_2019}, (3) yield optimization for a borylation reaction \citep{stevens_advancing_2022}, and (4) yield optimization for deoxyfluorination reaction \citep{nielsen_deoxyfluorination_2018}. 
Since it has been well-demonstrated that these problems can be effectively modeled by regression approaches \citep{zahrt_prediction_2019, ahneman_predicting_2018, sandfort_structure-based_2020}, we trained a random forest regressor on each dataset, which was used as the ground truth for all experiments \citep{hase_olympus_2021}.
To mitigate the known bias of \gls{hte} benchmarks towards high-outcome experiments \citep{strieth-kalthoff_machine_2022, beker_machine_2022}, we augment each dataset with a chemically sensible search space expansion workflow.
Notably, the random forest models were only trained on the originally published data, as it was used within the expansion workflow (see below).
In the following, the workflow and details of the datasets are described.

\subsubsection{Chemically sensible search space expansion workflow} \label{subsubsec: augmentation}

To mitigate the high number of high-outcome experiments (the respective search spaces were rationally designed by expert chemists), we augment them with more negative examples to make them more relevant to real-world optimization campaigns.
New substrates are generated by mutating the originally reported substrates via the STONED algorithm \citep{nigam_beyond_2021}.
In a first filtering step, new substrates were removed if they had a Tanimoto similarity to the original substrate smaller than $0.75$ ($0.6$ for the borylation reaction to obtain a reasonable number of additional substrates) or if they did not possess the functional groups required for the reaction.
To ensure that the benchmark is augmented with negative examples, random forest models are fitted to the original benchmarks (see above).
The mean absolute errors (MAEs), root-mean-square errors (RMSEs) and $R^2$ score ($R^2$), Spearman's rank correlation coefficient (Spearman's $\rho$) of the random forest regressors fitted to and evaluated on the original datasets are shown in \cref{tab:MAE}.
In addition, to evaluate the predictive utility of the random forest regressors, we perform 5-fold cross validation on the original benchmark. The MAE, RMSE, $R^2$ and Spearman's $\rho$ of the 5-fold cross validation are reported in \cref{tab:CV}. 
Even though the predictive performance on the CV does not achieve a high Spearman's rank coefficient, the comparably low MAEs and RMSEs, as well as high $R^2$ values suggest that they are a reasonable oracle.
Newly generated substrates were incorporated if the average reaction outcome over all reported reaction conditions is below a defined threshold.
The chosen thresholds are $1.0\%$ for the Pd-catalyzed carbon-heterocoupling, $0.7$\,kcal/mol for the \textit{N},\textit{S}-acetal formation, $12\%$ for the borylation reaction, and $5\%$ for the deoxyfluorination reaction.
If a substrate passed these filters, the reactions with all different reported conditions were added, with reaction outcomes being taken from as predicted from the random forest oracle.

\begin{table}[h]
    \centering
    \caption{MAE, RMSE, $R^2$, and Spearman's $\rho$ of random forest regressors fitted to and evaluated on the original benchmark problems.}
    \begin{tabularx}{\textwidth}{lXXXX}\toprule
           Benchmark problem & MAE & RMSE & $R^2$ & Spearman's $\rho$\\\midrule
Pd-catalyzed coupling    & $3.16 \times 10^{-3}$ & $8.75 \times 10^{-3}$ & $0.966$ & $0.898$\\
\textit{N},\textit{S}-acetal formation & $4.95 \times 10^{-2}$ \text{kcal/mol}& $7.39 \times 10^{-2}$ \text{kcal/mol} & $0.989$ & $0.994$\\
Borylation & $3.62 \times 10^{-2}$ & $4.92 \times 10^{-2}$ & $0.966$ & $0.987$\\
Deoxyfluorination & $2.13 \times 10^{-2}$ & $3.38 \times 10^{-2}$ & $0.986$ & $0.993$\\\bottomrule
    \end{tabularx}
    \label{tab:MAE}
\end{table}

\begin{table}[h]
    \centering
    \caption{MAE, RMSE, $R^2$, and Spearman's rank correlation coefficient with their standard errors of random forest regressors in a 5-fold cross validation on the original benchmark problems.}
    \begin{tabularx}{\textwidth}{lXXXX}\toprule
           Benchmark problem & MAE & RMSE & $R^2$ & Spearman's $\rho$\\\midrule
Pd-catalyzed coupling    & $(9.3 \pm 0.7) \times 10^{-3}$ & $(2.44 \pm 0.18) \times 10^{-2}$ & $0.73 \pm 0.03$ & $0.429 \pm 0.007$\\
\textit{N},\textit{S}-acetal formation & $(1.43 \pm 0.07) \times 10^{-1}$\,\text{kcal/mol} & $(2.11 \pm 0.10) \times 10^{-1}$\,\text{kcal/mol} & $0.908 \pm 0.010$ & $0.474 \pm 0.007$\\
Borylation & $(1.04 \pm 0.03) \times 10^{-1}$ & $(1.39 \pm 0.04) \times 10^{-1}$ & $0.729 \pm 0.013$ & $0.425 \pm 0.009$\\
Deoxyfluorination & $(5.96 \pm 0.14) \times 10^{-2}$ & $(8.42 \pm 0.15) \times 10^{-2}$ & $0.913 \pm 0.004$ & $0.478 \pm 0.003$\\\bottomrule
    \end{tabularx}
    \label{tab:CV}
\end{table}
\newpage
\subsubsection{Details on the expanded datasets} \label{subsubsec:dataset_details}

\textbf{Pd-catalyzed carbon-heterocoupling}

\begin{figure}[h]
    \centering
    \includegraphics[]{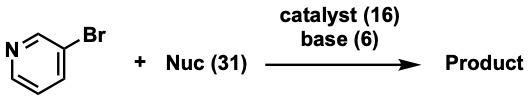}
    \caption{Reaction diagram of the Pd-catalyzed carbon-heterocoupling, where 3-bromopyridine reacts with a nucleophile. Reaction conditions include a catalyst and a base. The numbers indicate the amount of different species in the expanded dataset.}
    \label{fig:Cernak_reaction_augmented}
\end{figure}

The expanded Pd-catalzyed carbon-heterocoupling dataset contains 31 different nucleophiles as substrates, up from 16 in the published dataset (see \cref{fig:Cernak_reaction_augmented}). 
Combined with the 96 reported reaction condition combinations, the expanded dataset consists of 2976 reactions, for which the conversion is reported.

\begin{figure}[h]
    \centering
    \includegraphics[]{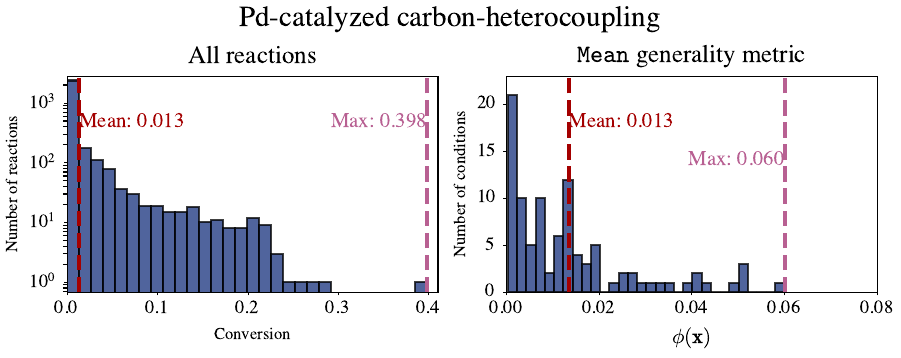}
    \includegraphics[]{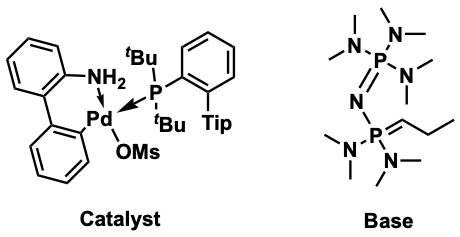}
    \caption{Top left: Distribution of the conversion for the Pd-catalyzed carbon-heterocoupling expanded dataset. Top right: Distribution of the average conversion for each catalyst-base combination for the Pd-catalyzed carbon-heterocoupling expanded dataset. Bottom: Catalyst-base combination with the highest average conversion in the expanded dataset. Tip = 2,4,6-triisopropylphenyl.}
    \label{fig:Cernak_EDA_augmented}
\end{figure}

Expansion decreased the average conversion across all reactions from $0.0205$ in the published dataset to $0.0134$, whereas the maximum conversion remained the same at $0.3981$ (see \cref{fig:Cernak_EDA_augmented}).
The average \gls{mean} conversion of each condition is decreased from $0.0205$ to $0.0134$, and the maximum \gls{mean} conversion of each condition is also decreased from $0.0760$ to $0.0600$ (see \cref{fig:Cernak_EDA_augmented}).
The catalyst-base combination with the highest \gls{mean} conversion is unaffected by the expansion and shown in \cref{fig:Cernak_EDA_augmented}.

\begin{figure}[h]
    \centering
    \includegraphics[]{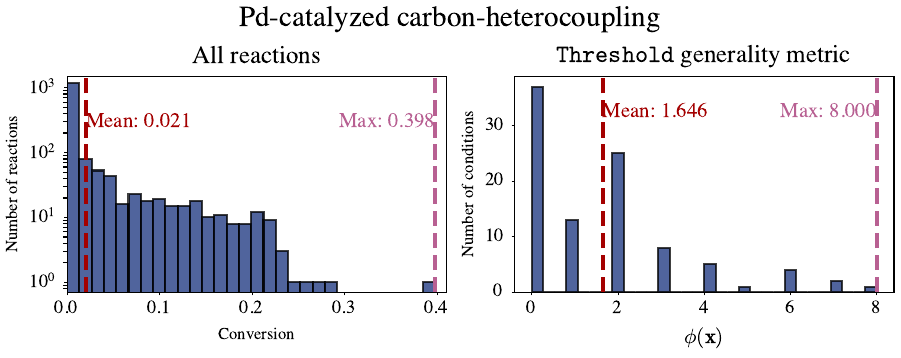}
    \caption{Left: Distribution of the conversion for the Pd-catalyzed carbon-heterocoupling expanded dataset. Right: Distribution of the number of substrates with a conversion above the specified threshold for each catalyst-base combination for the Pd-catalyzed carbon-heterocoupling expanded dataset.}
    \label{fig:Cernak_EDA_frac_augmented}
\end{figure}

With respect to the \gls{threshold} generality metric, the chosen threshold was $0.0750$.
The average number of substrates with a conversion above this threshold are $1.646$, while the maximum number of substrates is $8$ (\cref{fig:Cernak_EDA_frac_augmented}).
The catalyst-base combination with the highest number of substrates with a conversion above the threshold is the same as shown in \cref{fig:Cernak_EDA_augmented}.

\begin{figure}[h]
    \centering
    \includegraphics[width=0.7\linewidth]{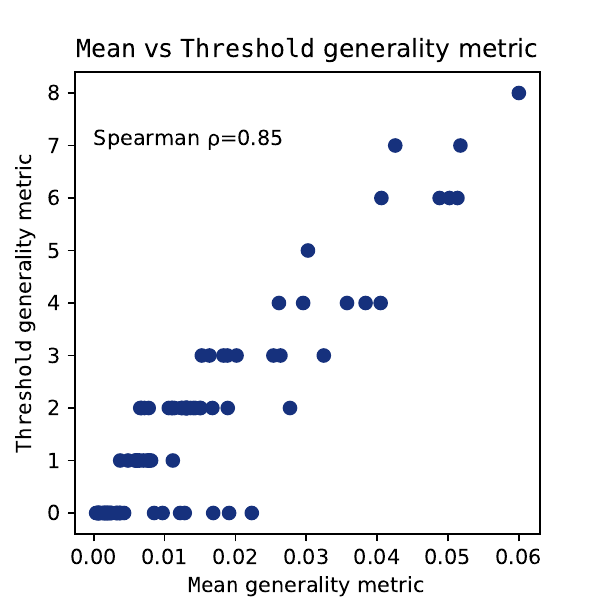}
    \caption{Distribution of the \gls{mean} \textit{vs.} \gls{threshold} generality metrics for the Pd-catalyzed carbon-heterocoupling expanded dataset.}
    \label{fig:Cernak_dataset_complexity}
\end{figure}

To estimate dataset complexity, we plot the \gls{mean} \textit{vs.} the \gls{threshold} generality metrics for each condition.
As shown in \cref{fig:Cernak_dataset_complexity}, these two metrics agree well with each other (Spearman's $\rho=0.85$).
Such an agreement indicates that conditions that give high conversions (\gls{mean} generality metric) do so for multiple substrates (\gls{threshold} generality metric), and that the \gls{mean} is not artificially inflated by outliers of specific substrates.
Additionally, the average span of conversions per conditions (\textit{i.e.}, maximum - minimum conversion for each condition) is $0.123\pm0.083$, significantly lower than the overall span of the dataset of $0.398$.
Lastly, we also hypothesized that the \gls{threshold} generality metric is hard to optimize (see above), because reactions that are only marginally below the set threshold are assigned the same score as reactions that are far below the threshold.
To quantify how different datasets might suffer from this effect, we investigate how many conditions have at least one reaction within $10\%$ of the threshold.
For this dataset, we thus investigate how many conditions have at least one reaction between $0.0675$ and $0.0750$ conversion.
We find that this is only the case for $18$ of $96$ conditions ($18.8\%$), showing that many conditions perform badly across all substrates.
Overall, these numbers indicate that the outcome of a reaction in the Pd-catalyzed carbon-heterocoupling dataset is mostly governed by the conditions, which mostly perform similarly across all considered substrates.

\newpage
\textbf{\textit{N},\textit{S}-acetal formation}

\begin{figure}[h]
    \centering
    \includegraphics[]{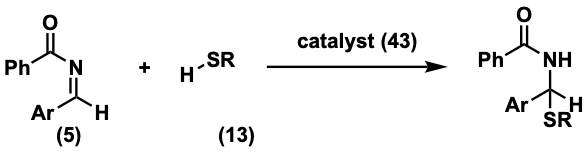}
    \caption{Reaction diagram of the \textit{N},\textit{S}-acetal formation, where an imine reacts with a thiol. Reaction conditions include a catalyst. The numbers indicate the amount of different species in the expanded dataset.}
    \label{fig:Denmark_reaction_augmented}
\end{figure}

The expanded \textit{N},\textit{S}-acetal formation formation dataset contains 13 thiols (up from 5 in the published dataset), and remains at 5 imines (see \cref{fig:Denmark_reaction_augmented}).
Combined with the 43 reported reaction conditions, the expanded dataset consists of 2795 reactions, for which $\Delta\Delta G^{\ddagger}$ is reported.

\begin{figure}[h]
    \centering
    \includegraphics[]{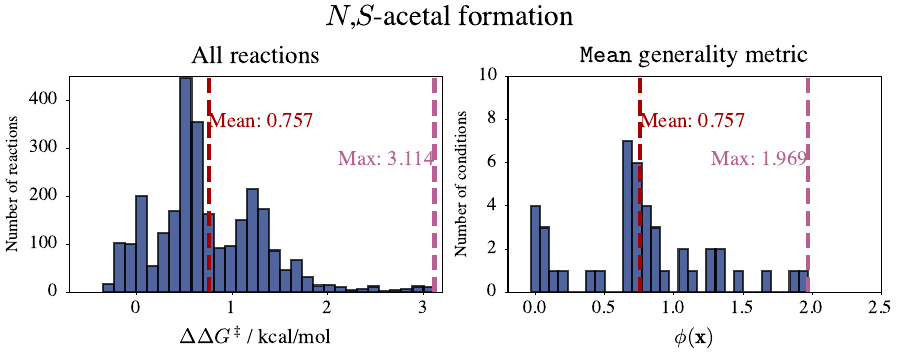}
    \includegraphics[]{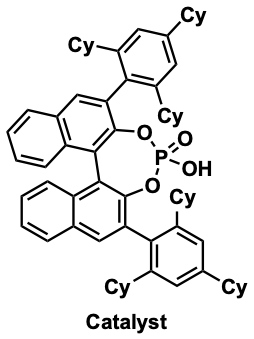}
    \caption{Top left: Distribution of $\Delta\Delta G^{\ddagger}$ for the \textit{N},\textit{S}-acetal formation expanded dataset. Top right: Distribution of the average $\Delta\Delta G^{\ddagger}$ for each catalyst for the \textit{N},\textit{S}-acetal formation expanded dataset. Bottom: Catalyst with the highest average $\Delta\Delta G^{\ddagger}$ expanded dataset. Cy = Cyclohexyl.}
    \label{fig:Denmark_EDA_augmented}
\end{figure}

Expansion decreased the average $\Delta\Delta G^{\ddagger}$ from $0.988$ \,kcal/mol in the published dataset to $0.757$ \,kcal/mol, whereas the maximum $\Delta\Delta G^{\ddagger}$ was slightly decreased from $3.135$\,kcal/mol to $3.114$\,kcal/mol (see  \cref{fig:Denmark_EDA_augmented}). 
This decrease is due to the fact that the expanded dataset only contains values that are taken as predicted by the random forest regressor.
Through expansion, the average \gls{mean} $\Delta\Delta G^{\ddagger}$ of each condition decreased from $0.988$\,kcal/mol to $0.757$\,kcal/mol, while the maximum \gls{mean} $\Delta\Delta G^{\ddagger}$ of all conditions decreased as well from $2.395$\,kcal/mol to $1.969$\,kcal/mol (see \cref{fig:Denmark_EDA_augmented}).
The catalyst with the highest \gls{mean} $\Delta\Delta G^{\ddagger}$ is unaffected by the expansion and shown in \cref{fig:Denmark_EDA_augmented}.

\begin{figure}[h]
    \centering
    \includegraphics[]{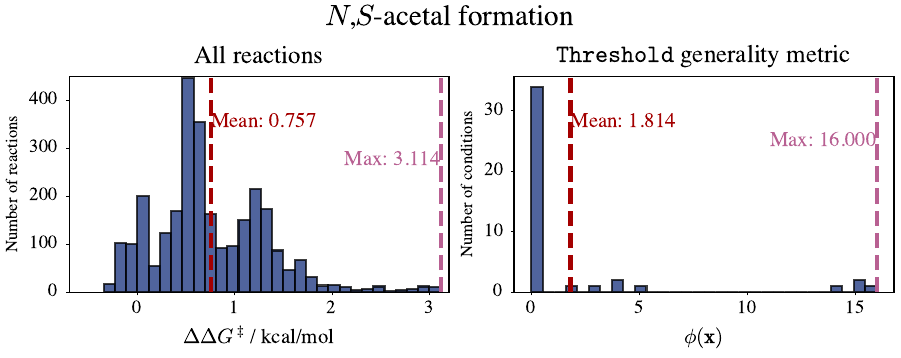}
    \caption{Left: Distribution of $\Delta\Delta G^{\ddagger}$ for the \textit{N},\textit{S}-acetal formation expanded dataset. Right: Distribution number of substrate combinations with a  $\Delta\Delta G^{\ddagger}$ above the specified threshold for each catalyst for the \textit{N},\textit{S}-acetal formation expanded dataset.}
    \label{fig:Denmark_EDA_frac_augmented}
\end{figure}

With respect to the \gls{threshold} generality metric, the chosen threshold was $2.0$\,kcal/mol.
The average number of substrate combinations (distinct imine/thiol combinations) with $\Delta\Delta G^{\ddagger}$ above this threshold are $1.814$, while the maximum number of substrate combinations is $16$ (\cref{fig:Denmark_EDA_frac_augmented}).
The catalyst with the highest number of substrate combinations with $\Delta\Delta G^{\ddagger}$ above the threshold is the same as shown in \cref{fig:Denmark_EDA_augmented}.

\begin{figure}[h]
    \centering
    \includegraphics[width=0.7\linewidth]{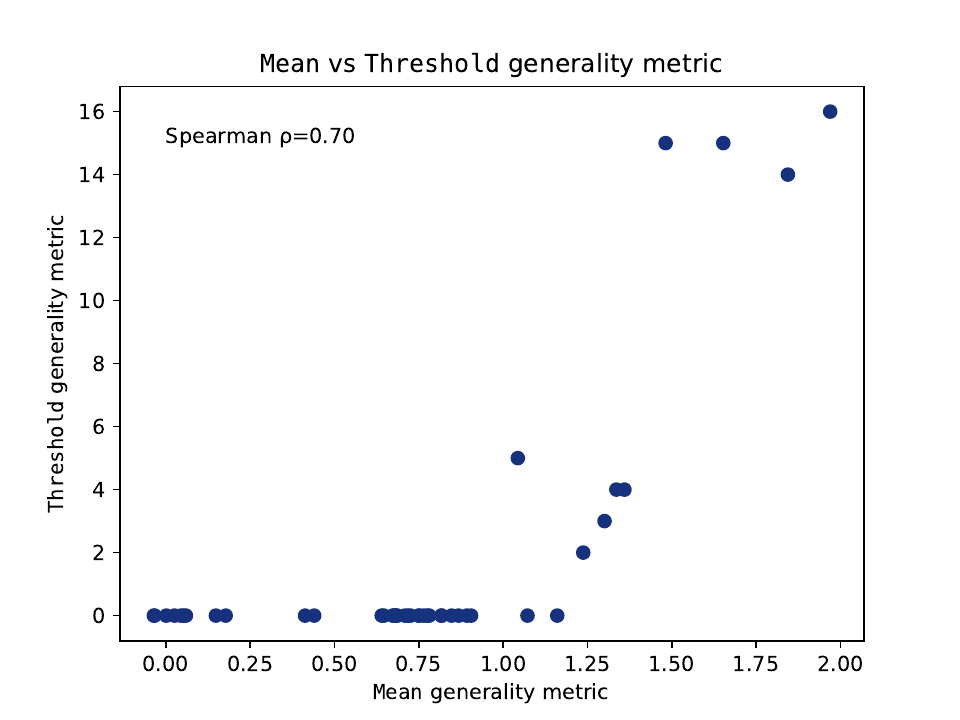}
    \caption{Distribution of the \gls{mean} \textit{vs.} \gls{threshold} generality metrics for the \textit{N},\textit{S}-acetal formation expanded dataset.}
    \label{fig:Denmark_dataset_complexity}
\end{figure}

To estimate dataset complexity, we plot the \gls{mean} \textit{vs.} the \gls{threshold} generality metrics for each condition.
As shown in \cref{fig:Denmark_dataset_complexity}, both metrics are dominated by four catalysts that achieve high scores in both generality metrics, while all other catalysts give significantly worse scores for both metrics.
Such an agreement indicates that conditions that give high $\Delta\Delta G^{\ddagger}$ (\gls{mean} generality metric) do so for multiple substrate combinations (\gls{threshold} generality metric), and that the \gls{mean} is not artificially inflated by outliers of specific substrate combinations.
Additionally, the average span of $\Delta\Delta G^{\ddagger}$ per condition (\textit{i.e.}, maximum -- minimum $\Delta\Delta G^{\ddagger}$ for each condition) is $1.02\pm0.41$\,kcal/mol, significantly lower than the overall span of the dataset of $3.46$\,kcal/mol.
Lastly, we also hypothesized that the \gls{threshold} generality metric is hard to optimize (see above), because reactions that are only marginally below the set threshold are assigned the same score as reactions that are far below the threshold.
To quantify how different datasets might suffer from this effect, we investigate how many conditions have at least one reaction within $10\%$ of the threshold.
For this dataset, we thus investigate how many conditions have at least one reaction with $\Delta\Delta G^{\ddagger}$ between $1.80$\,kcal/mol and $2.00$\,kcal/mol.
We find that this is only the case for $9$ of $43$ conditions ($20.9\%$), showing that many conditions perform badly across all substrate combinations.
Overall, these numbers indicate that the outcome of a reaction in the \textit{N},\textit{S}-acetal formation dataset is mostly governed by the catalysts, which mostly perform similarly across all considered substrate combinations.

\newpage
\textbf{Borylation reaction}

\begin{figure}[h]
    \centering
    \includegraphics[]{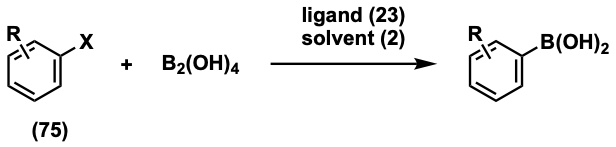}
    \caption{Reaction diagram of the borylation reaction, where an aryl electrophile is borylated via a nickel catalyst. Reaction conditions include a ligand, and a solvent. The numbers indicate the amount of different species in the expanded dataset.}
    \label{fig:Borylation_reaction_augmented}
\end{figure}

The expanded borylation dataset consists of 75 different aryl electrophiles, up from 33 in the originally published dataset (see \cref{fig:Borylation_reaction_augmented}). 
Combined with the 46 reported reaction condition combinations, the augmented dataset consists of 3450 reactions, for which the yield is reported.

\begin{figure}[h]
    \centering
    \includegraphics[]{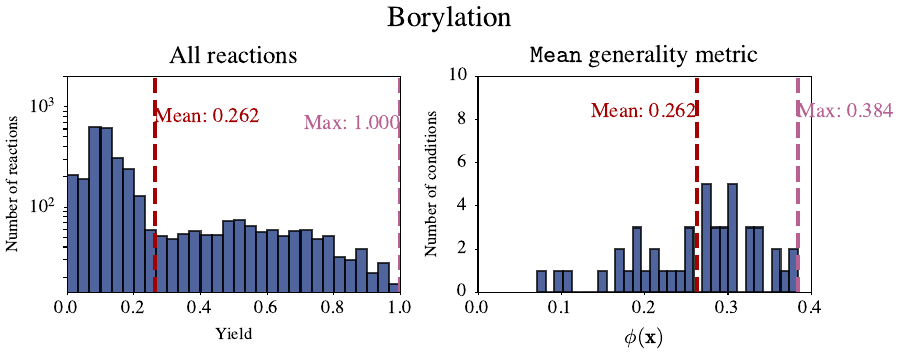}
    \includegraphics[]{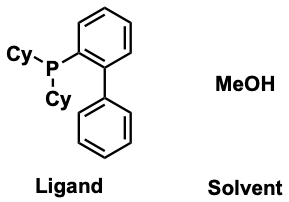}
    \caption{Top left: Distribution of the yield for the borylation reaction expanded dataset. Top right: Distribution of the average yield for each ligand-solvent combination for the borylation reaction expanded dataset. Bottom: Ligand-solvent combination with the highest average yield expanded dataset. Cy = Cyclohexyl, Me = Methyl.}
    \label{fig:Borylation_EDA_augmented}
\end{figure}

Expansion decreased the average yield from $0.455$ to $0.262$, whereas the maximum yield remained the same at $1.000$ (see \cref{fig:Borylation_EDA_augmented}).
The average \gls{mean} yield of each condition is decreased from $0.455$ to $0.262$, and the maximum \gls{mean} yield of each condition is also decreased from $0.654$ to $0.384$ (see \cref{fig:Borylation_EDA_augmented}).
The ligand-solvent combination with the highest \gls{mean} yield is unaffected by dataset and expansion and shown in \cref{fig:Borylation_EDA_augmented}.

\begin{figure}[h]
    \centering
    \includegraphics[]{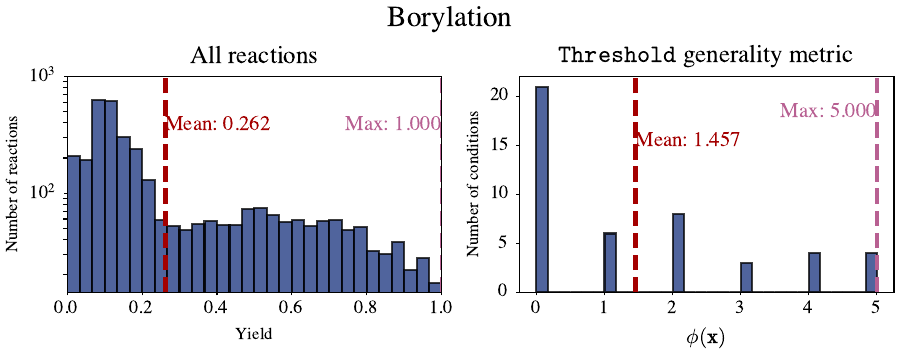}
    \caption{Left: Distribution of the yield for the borylation reaction expanded dataset. Right: Distribution of the number of substrates with a yield above the specified threshold for each ligand-solvent combination for the borylation reaction in the expanded dataset.}
    \label{fig:Borylation_EDA_frac_augmented}
\end{figure}

With respect to the \gls{threshold} generality metric, the chosen threshold was $0.90$.
The average number of substrates with a yield above this threshold are $1.457$, while the maximum number of substrates is $5$ (\cref{fig:Borylation_EDA_frac_augmented}).
Several ligand-solvent combinations provide the highest number of substrates with a yield above the threshold, one of them is shown in \cref{fig:Borylation_EDA_augmented}.
The ligand-solvent combinations are unaffected by the expansion.

\begin{figure}[h]
    \centering
    \includegraphics[width=0.7\linewidth]{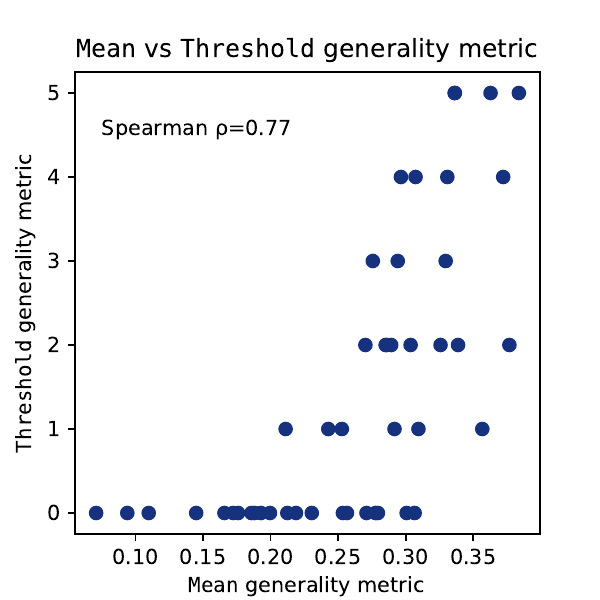}
    \caption{Distribution of the \gls{mean} \textit{vs.} \gls{threshold} generality metrics for the borylation expanded dataset.}
    \label{fig:Borylation_dataset_complexity}
\end{figure}

To estimate dataset complexity, we plot the \gls{mean} \textit{vs.} the \gls{threshold} generality metrics for each condition.
As shown in \cref{fig:Borylation_dataset_complexity}, a trend is generally recognizable, as conditions often give high scores for both generality metrics.
However, in contrast to previous datasets, there are no few conditions that give significantly higher scores than the rest.
Instead, multiple conditions perform well on both metrics, even achieving the maximum score on the \gls{threshold} generality metric.
Additionally, some conditions achieve close to the best score on the \gls{mean} generality metric, but perform significantly worse on the \gls{threshold} metric.
These results suggest that the \gls{mean} generality metric can be susceptible to outliers, a caveat that chemists have to be aware of when designing their scope of considered substrates.
Additionally, the average span of yield per conditions, (\textit{i.e.}, maximum -- minimum yield for each condition) is $0.874\pm0.121$, indicating that the reaction outcome is not goverend by the conditions, but highly influenced to the precise combination of conditions/substrate combination.
Lastly, we also hypothesized that the \gls{threshold} generality metric is hard to optimize (see above), because reactions that are only marginally below the set threshold are assigned the same score as reactions that are far below the threshold.
To quantify how different datasets might suffer from this effect, we investigate how many conditions have at least one reaction within $10\%$ of the threshold.
For this dataset, we thus investigate how many conditions have at least one reaction with a yield between $0.81$ and $0.90$.
We find that this is the case for $34$ of $46$ conditions ($73.9\%$), showing that many conditions perform well for at least some substrates.
Overall, these numbers indicate that the outcome of a reaction in the borylation dataset is governed by the precise conditions/substrate combination.
As multiple conditions also achieve the maximum score on the \gls{threshold} generality metric, the borylation dataset is challenging for generality-oriented optimization.

\newpage
\textbf{Deoxyfluorination reaction}

\begin{figure}[h]
    \centering
    \includegraphics[]{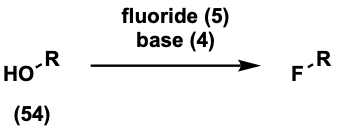}
    \caption{Reaction diagram of the deoxyfluorination reaction, where an alcohol is converted to the corresponding fluoride. Reaction conditions include a fluoride source and a base. The numbers indicate the amount of different species in the expanded dataset.}
    \label{fig:Deoxyfluorination_reaction_augmented}
\end{figure}

The expanded deoxyfluorination dataset consists of 54 different alcohols, up from 37 in the originally published dataset (see \cref{fig:Deoxyfluorination_reaction_augmented}). Combined with the 20 reported reaction condition combinations, the augmented dataset consists of 1080 reactions, for which the yield is reported.

\begin{figure}[h]
    \centering
    \includegraphics[]{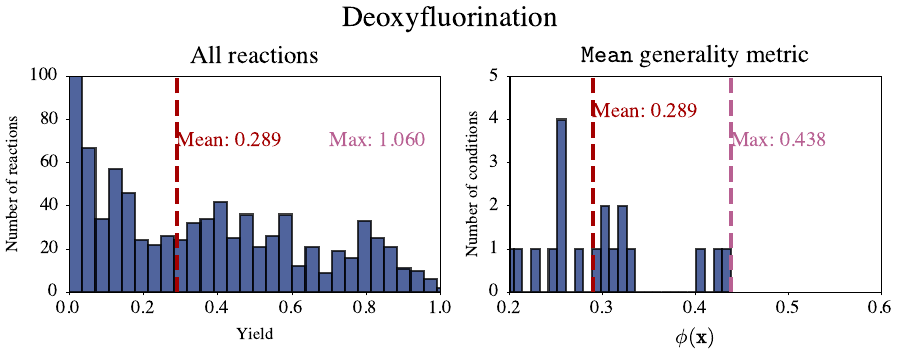}
    \includegraphics[]{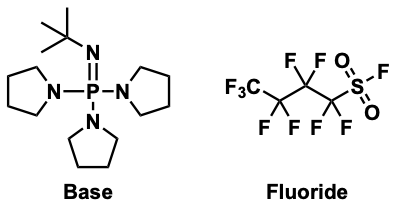}
    \caption{Top left: Distribution of the yield for the deoxyfluorination reaction expanded dataset. Top right: Distribution of the average yield for each fluoride-base combination for the deoxyfluorination reaction expanded dataset. Bottom: Fluoride-base combination with the highest average yield expanded dataset.}
    \label{fig:Deoxyfluorination_EDA_augmented}
\end{figure}

Expansion decreased the average yield from $0.404$ to $0.289$, whereas the maximum yield remained the same at $1.006$ (see \cref{fig:Deoxyfluorination_EDA_augmented}).
The yield larger than $1.00$ is contained in the originally published dataset.
The average \gls{mean} yield of each condition is decreased from $0.404$ to $0.289$, and the maximum \gls{mean} yield of each condition is also decreased from $0.572$ to $0.438$ (see \cref{fig:Deoxyfluorination_EDA_augmented}).
The fluoride-base combination with the highest \gls{mean} yield is unaffected by expansion and shown in \cref{fig:Deoxyfluorination_EDA_augmented}.

\begin{figure}[h]
    \centering
    \includegraphics[]{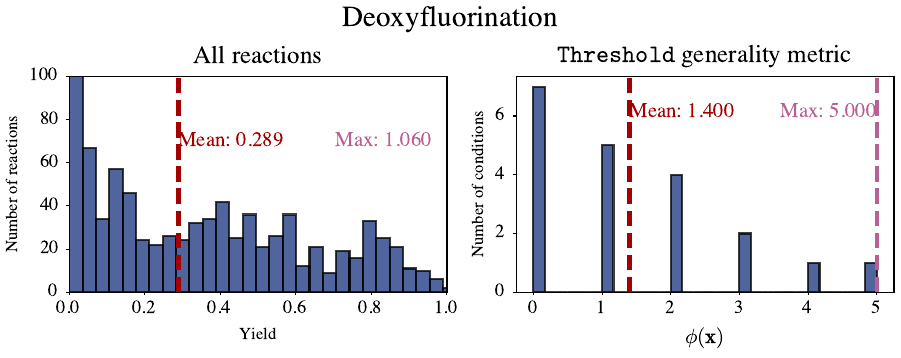}
    \includegraphics[]{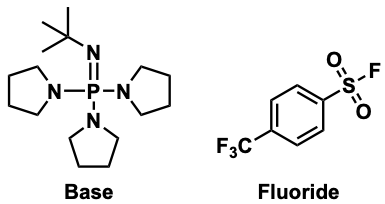}
    \caption{Top left: Distribution of the yield for the deoxyfluorination reaction in the expanded dataset. Top right: Distribution of the number of substrates with a yield above the specified threshold for each fluoride-base combination for the deoxyfluorination reaction in the expanded dataset. Bottom: Fluoride-base combination with the highest number of substrate with a yield above the threshold in the expanded dataset.}
    \label{fig:Deoxyfluorination_EDA_frac_augmented}
\end{figure}

With respect to the \gls{threshold} generality metric, the chosen threshold was $0.90$.
The average number of substrates with a yield above this threshold are $1.400$, while the maximum number of substrates is $5$ (\cref{fig:Deoxyfluorination_EDA_frac_augmented}).
The fluoride-base combination with the highest number of substrates with a yield above the threshold is also unaffected by expansion and shown in \cref{fig:Deoxyfluorination_EDA_frac_augmented}.

\begin{figure}[h]
    \centering
    \includegraphics[width=0.7\linewidth]{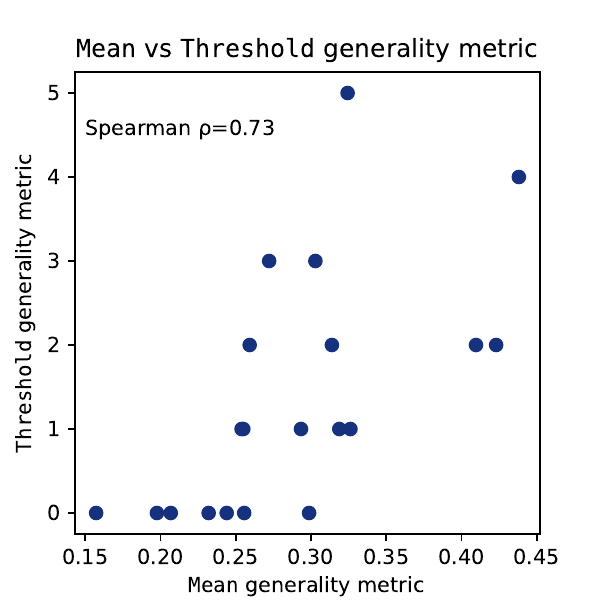}
    \caption{Distribution of the \gls{mean} \textit{vs.} \gls{threshold} generality metrics for the deoxyfluorination expanded dataset.}
    \label{fig:Deoxyfluorination_dataset_complexity}
\end{figure}

To estimate dataset complexity, we plot the \gls{mean} \textit{vs.} the \gls{threshold} generality metrics for each condition.
As shown in \cref{fig:Deoxyfluorination_dataset_complexity}, a trend is generally recognizable, as conditions often give high scores for both generality metrics.
However, conditions that achieve the highest score on the \gls{mean} generality metric do not achieve the highest score on the \gls{threshold} generality metric.
This result suggest that the \gls{mean} generality metric can be susceptible to outliers, a caveat that chemists have to be aware of when designing their scope of considered substrates.
Additionally, the average span of yield per conditions (\textit{i.e.}, maximum -- minimum yield for each condition) is $0.894\pm0.124$, indicating that the reaction outcome is not governed by the conditions, but highly influenced to the precise combination of conditions/substrate combination.
Lastly, we also hypothesized that the \gls{threshold} generality metric is hard to optimize (see above), because reactions that are only marginally below the set threshold are assigned the same score as reactions that are far below the threshold.
To quantify how different datasets might suffer from this effect, we investigate how many conditions have at least one reaction within $10\%$ of the threshold.
For this dataset, we thus investigate how many conditions have a reaction with a yield between $0.81$ and $0.90$.
We find that this is the case for $16$ of $20$ conditions ($80.0\%$), showing that many conditions perform well for at least some substrates.
Overall, these numbers indicate that the outcome of a reaction in the deoxyfluorination dataset is governed by the precise conditions/substrate combination, making this dataset challenging for generality-oriented optimization.

\newpage
\subsection{Details on benchmarking for BO under partial monitoring} \label{subsec:BO_benchmark}

\subsubsection{Details on \gls{currybo} optimization algorithms}
To identify whether \gls{bo} under partial monitoring, as described above, can efficiently identify general optima, we conducted several benchmarking runs on the described benchmark problems.
On each dataset, we perform benchmarking for multiple optimization strategies, as listed in \cref{tab:acquisition_strategies_all}.
In each optimization campaign, we used a single-task GP regressor, as implemented in \textit{GPyTorch} \citep{gardner_gpytorch_2018}, with a TanimotoKernel as implemented in \textit{Gauche} \citep{griffiths_gauche_2023}.
Molecules were represented using Morgan Fingerprints \citep{morgan_generation_1965} with 1024 bits and a radius of 2. 
Fingerprints were generated using RDKit \citep{landrum_rdkit_2023}. 
It is notable that, while such a representation was chosen due to its suitability for broad chemical spaces, more specific representations such as descriptors might be able to improve the optimization performance.

The acquisition policies were benchmarked on all benchmark problems with differently sampled substrates for each optimization run.
For each benchmark, we selected the train set randomly, consisting of twelve nucleophiles in the Pd-catalyzed carbon-heterocoupling benchmark, three imines and three thiols in the \textit{N},\textit{S}-acetal formation benchmark, 25 alcohols in the deoxyfluorination reaction, and twenty aryl halides in the borylation reaction.
For each, 30 independent optimization campaigns were performed, differing in substrates in the train set and initially measured conditions (always starting from one measurement).
The generality of the proposed general conditions at each step during the optimization is shown.

Each independent run was performed on one CPU with 4\,GB memory per CPU on an HPC cluster.
The runtimes for independent runs with each acquisition strategy and dataset are given in \cref{tab:runtime}.
Storing the results of one independent run takes approximately 6\,MB of storage space.
Note that for each acquisition strategy, the runtime is on the same order of magnitude, independent of $\alpha_x$ and $\alpha_w$ and also whether the \gls{mean} generality metric or \gls{threshold} generality metric is evaluated.

\newcolumntype{Y}{>{\centering\arraybackslash}X}
\begin{table*}[t]
    \footnotesize
    \centering
    \caption{
    Runtime for different types of acquisition strategies on the tested real-world benchmark problems. The mean (and standard deviation in brackets) over all independent runs is given.
    }

    \vspace{0.5em}
    
    \renewcommand{\arraystretch}{1.3}

    \begin{tabularx}{\textwidth}{p{0.25\textwidth}Y Y Y}\toprule
        \textbf{Benchmark} & \textbf{One-step lookahead} & \multicolumn{2}{c}{\textbf{Two-step lookahead}} \\ \cmidrule(lr){3-4}
         & Sequential / h & Sequential / h & Joint / h \\ \midrule
        Pd-catalyzed carbon-heterocoupling & $0.11\pm0.01$ & $35\pm6$ & $170\pm3$ \\
        \textit{N},\textit{S}-acetal formation & $0.047\pm0.006$ & $5.5\pm0.8$ & $34\pm1$\\
        Borylation & $0.11\pm0.02$ & $18\pm2$ & $111\pm2$\\
        Deoxyfluorination & $0.069\pm0.008$ & $6\pm1$ & $38\pm1$\\
        \bottomrule
    \end{tabularx}
    \label{tab:runtime}
\end{table*}

\newpage
\subsection{Additional Results} \label{subsec:add_results}

\subsubsection{Additional Results on Grid Search for Dataset Analysis} \label{subsubsec:add_results_data-analysis}

\begin{figure*}[t]
    \centering
    \includegraphics[width=\linewidth]{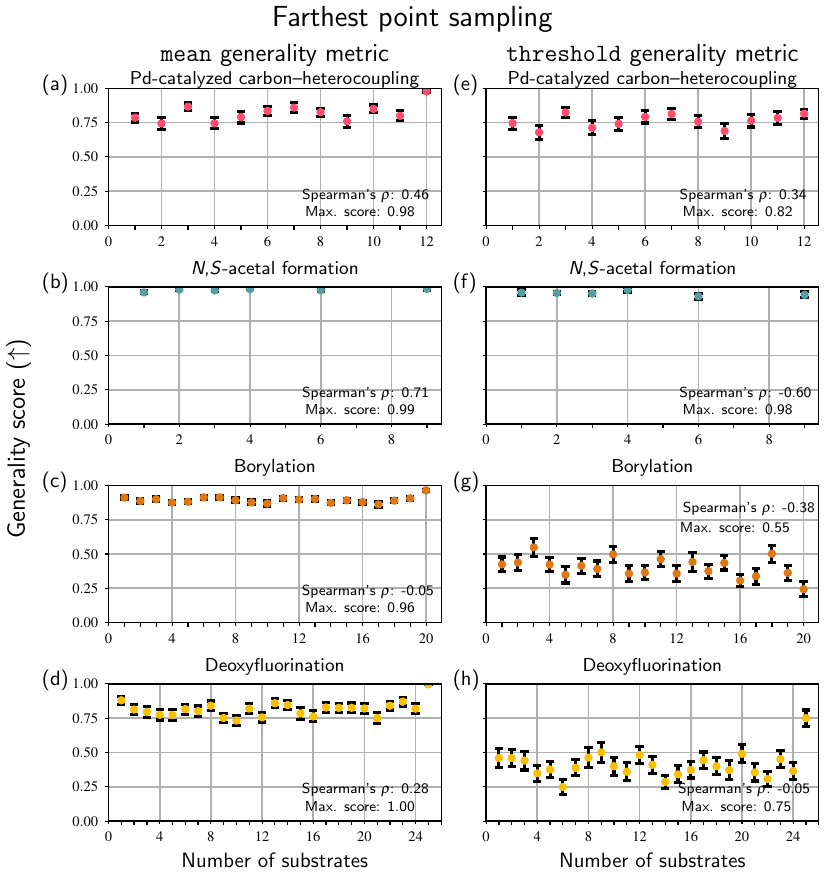}
    \vspace{-1em}
    \caption{
        Test-set generality score given by the \gls{mean} (left) and \gls{threshold} (right) metrics, as determined by exhaustive grid search for the four datasets. Train/test splits were performed using farthest point sampling. Average and standard error are calculated from the 30 different train/test substrate splits.
    }
    \label{fig:add_results_dataset_analysis_fps}
\end{figure*}

\begin{figure*}[t]
    \centering
    \includegraphics[width=\linewidth]{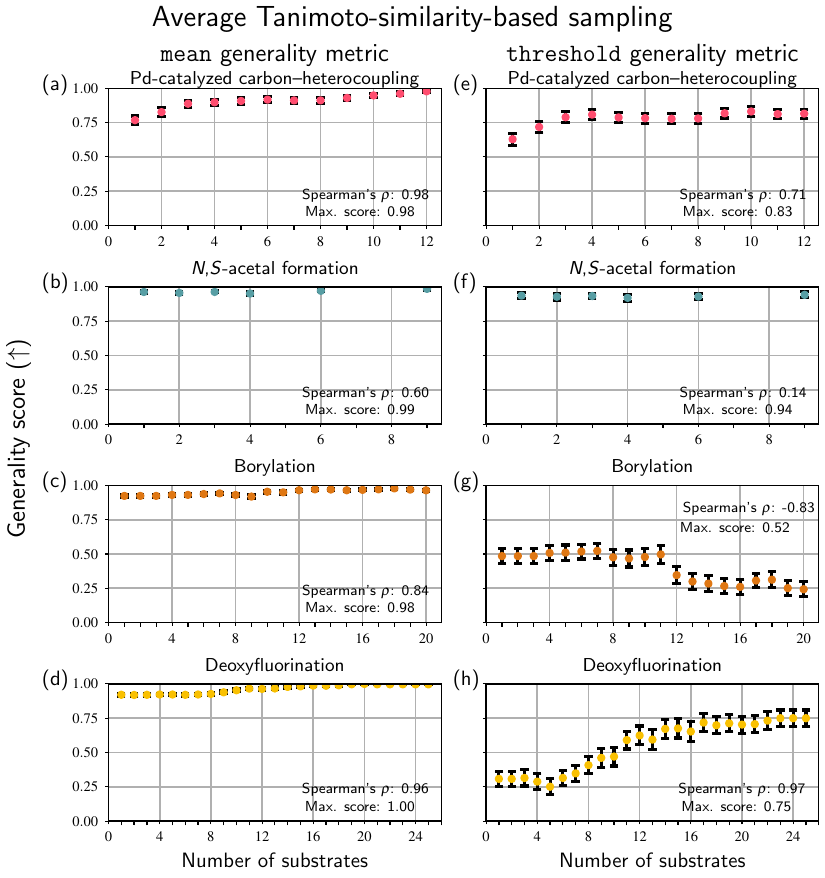}
    \vspace{-1em}
    \caption{
        Test-set generality score given by the \gls{mean} (left) and \gls{threshold} (right) metrics, as determined by exhaustive grid search for the four datasets. Train set was selected using the highest average Tanimoto similarity to unselected substrates. Average and standard error are calculated from the 30 different train/test substrate splits.
    }
    \label{fig:add_results_dataset_analysis_smart}
\end{figure*}

In the main text, we analyzed the generality score of conditions, identified with a varying number of train substrates, to an unseen test set.
As outlined, there we considered a random split between train and test substrates.
We further considered two alternative splitting strategies: (1) Split substrates based on farthest point sampling, and (2) iteratively add the substrate to the train set with the highest average Tanimoto similarity to all yet unadded substrates.
Both strategies aim to broadly cover the space of substrates, emulating how a chemist might select substrates for a generality-oriented optimization campaign.
Based on the new train/test splits, we perform the same analysis as outlined above, with results shown in \cref{fig:add_results_dataset_analysis_fps,fig:add_results_dataset_analysis_smart}, respectively.

As demonstrated in \cref{fig:add_results_dataset_analysis_fps,fig:add_results_dataset_analysis_smart} on the \gls{mean} generality metric, high generality scores can be achieved on all datasets.
For a majority of datasets, optimizing over multiple substrates leads to more general conditions, as evidenced by Spearman's $\rho > 0$.
The only exception is the borylation dataset, where the reaction outcome is highly dependent on the precise conditions/substrate combination (see above).
Nonetheless, very high theoretical generality scores are observed across the board through multi-substrate optimization.

Considering the \gls{threshold} generality metric, very high generality scores are again observed for the Pd-catalyzed carbon-heterocoupling and the \textit{N},\textit{S}-acetal formation datasets, where the reaction outcome is mostly determined by the conditions.
On the borylation and deoxyflourination datasets, lower generality scores are observed, consistent with the observations presented in the main text.
This low score, and the observation that multi-substrate optimization does not lead to big improvements in generality score on these datasets, can be rationalized by their reaction characteristics.
In those datasets, the outcome of a reaction is highly dependent on the precise conditions/substrate combination, which renders identifying general conditions difficult.
This intricacy makes the most general conditions highly dependent on the chosen train substrates, and attention should be paid to substrate selection in such cases.

Nonetheless, the presented results herein agree with the one in the main text and underline that the observed trends are valid for other train/test split strategies: for a majority of datasets, performing multi-substrate optimization can lead to more general conditions than single-substrate optimization.
It is particularly notable that farthest point sampling did not outperform other sampling techniques that are not design to maximize chemical space coverage (\textit{e.g.}, random), as this strategy is commonly used in the literature to select chemicals to broadly cover chemical space \citep{henle_development_2020, gensch_comprehensive_2022, gensch_design_2022, schnitzer_machine_2024}.

\subsubsection{Optimization Results for each Dataset}\label{subsubsec:add_results_augmented}

In the main text, we investigated which factors determine successful generality-oriented optimizations and showed the \gls{gap} value averaged over all datasets.
However, given their different characteristics outlined above, further insight can be gained by analyzing the datasets independently.

\textbf{Comparing sequential and joint acquisition strategies}

\begin{figure*}[t]
    \centering
    \includegraphics[width=\linewidth]{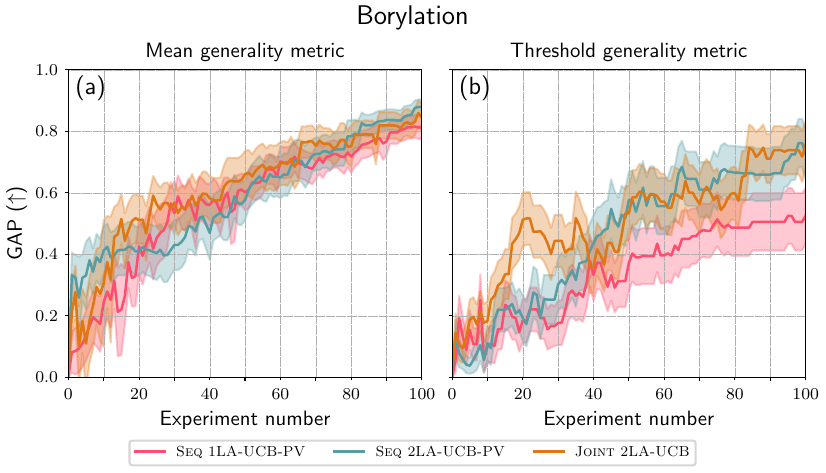}
    \vspace{-1em}
    \caption{
        Optimization trajectories for the sequential one-step lookahead (\gls{seqonelaucbpv}), sequential two-step lookahead (\gls{seqtwolaucbpv}) and joint two-step lookahead (\gls{jointtwolaucb}) acquisition strategies on the \gls{mean} (a) and \gls{threshold} (b) generality metrics on the borylation dataset.
    }
    \label{fig:add_results_strategies_borylation}
\end{figure*}

\begin{figure*}[t]
    \centering
    \includegraphics[width=\linewidth]{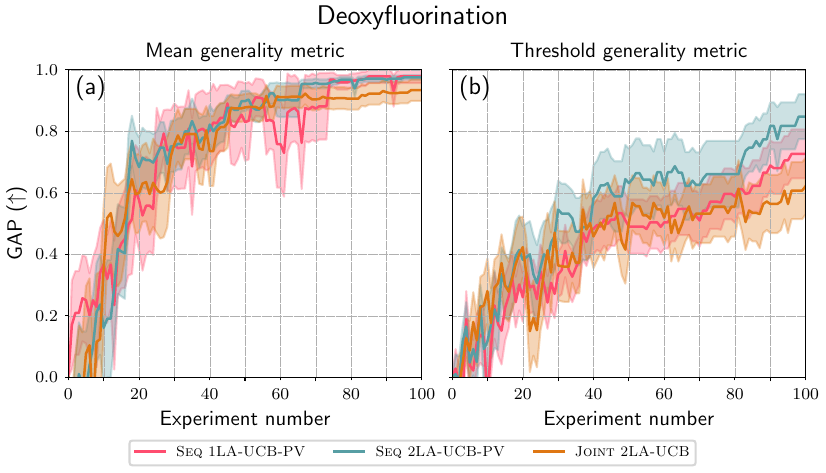}
    \vspace{-1em}
    \caption{
        Optimization trajectories for the sequential one-step lookahead (\gls{seqonelaucbpv}), sequential two-step lookahead (\gls{seqtwolaucbpv}) and joint two-step lookahead (\gls{jointtwolaucb}) acquisition strategies on the \gls{mean} (a) and \gls{threshold} (b) generality metrics on the deoxyfluorination dataset.
    }
    \label{fig:add_results_strategies_deoxyfluorination}
\end{figure*}

Similarly to \cref{fig:strategies}, we investigate the optimization of a \gls{seqonela}, \gls{seqtwola} and \gls{jointtwola} acquisition strategy on each of the four datasets.
As outlined in the main text, \gls{seqtwolaucbpv} achieves similarly or slightly higher \gls{gap} values than \gls{jointtwolaucb} for the two metrics.
This trend is observable across all the different datasets, suggesting that decoupling conditions and substrate selection is a valid assumption regardless of dataset characteristics.
Comparing the performance of \gls{seqtwolaucbpv} and \gls{seqonelaucbpv} across datasets, we find that the difference is most pronounced for the \gls{threshold} generality metric and is largely driven by the borylation and deoxyfluorination datasets (see \cref{fig:add_results_strategies_borylation,fig:add_results_strategies_deoxyfluorination}, respectively).
Given these are the datasets where optimization is expected to be harder (see above), these results demonstrate the value of a two-step lookahead acquisition strategy for these cases.

\begin{figure*}[t]
    \centering
    \includegraphics[width=\linewidth]{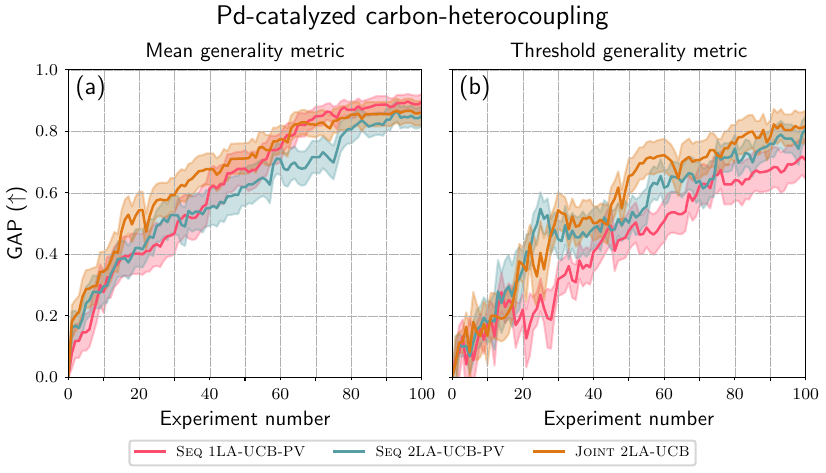}
    \vspace{-1em}
    \caption{
        Optimization trajectories for the sequential one-step lookahead (\gls{seqonelaucbpv}), sequential two-step lookahead (\gls{seqtwolaucbpv}) and joint two-step lookahead (\gls{jointtwolaucb}) acquisition strategies on the \gls{mean} (a) and \gls{threshold} (b) generality metrics on the Pd-catalyzed carbon-heterocoupling dataset.
    }
    \label{fig:add_results_strategies_Cernak}
\end{figure*}

\begin{figure*}[t]
    \centering
    \includegraphics[width=\linewidth]{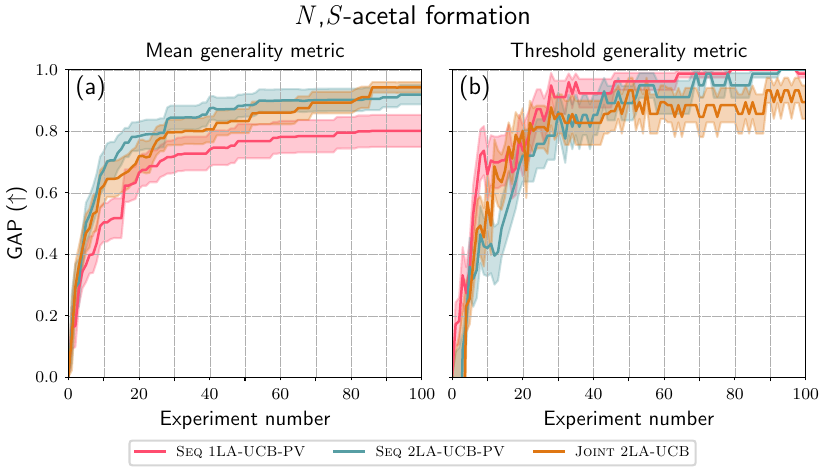}
    \vspace{-1em}
    \caption{
        Optimization trajectories for the sequential one-step lookahead (\gls{seqonelaucbpv}), sequential two-step lookahead (\gls{seqtwolaucbpv}) and joint two-step lookahead (\gls{jointtwolaucb}) acquisition strategies on the \gls{mean} (a) and \gls{threshold} (b) generality metrics on the \textit{N},\textit{S}-acetal formation dataset.
    }
    \label{fig:add_results_strategies_Denmark}
\end{figure*}

\textbf{Influence of Substrate Acquisition Function}

\begin{figure*}[t]
    \centering
    \includegraphics[width=\linewidth]{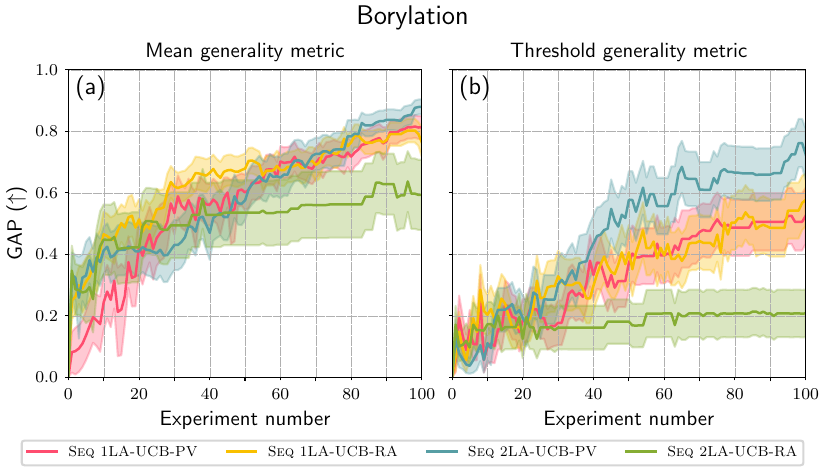}
    \vspace{-1em}
    \caption{
        Optimization trajectories for the \gls{seqonela} and \gls{seqtwola} acquisition strategies with \gls{pvfunc} and \gls{rafunc} substrate acquisition functions, on the \gls{mean} (a) and \gls{threshold} (b) generality metrics on the borylation dataset.
    }
    \label{fig:add_results_task-selection_borylation}
\end{figure*}

\begin{figure*}[t]
    \centering
    \includegraphics[width=\linewidth]{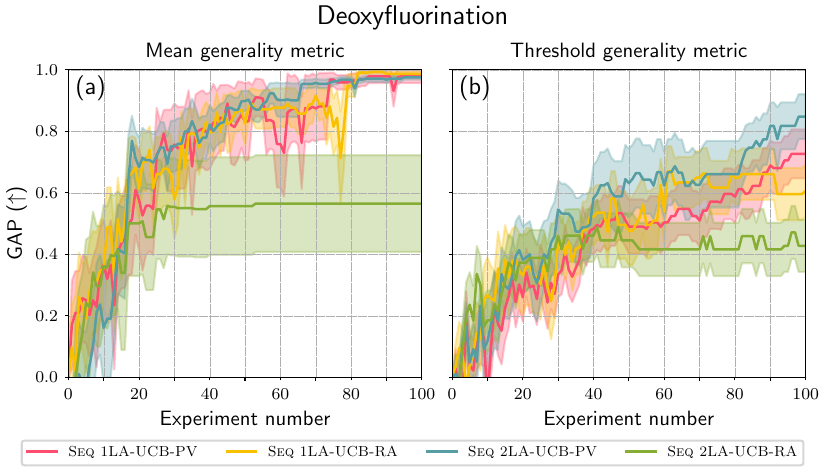}
    \vspace{-1em}
    \caption{
        Optimization trajectories for the \gls{seqonela} and \gls{seqtwola} acquisition strategies with \gls{pvfunc} and \gls{rafunc} substrate acquisition functions, on the \gls{mean} (a) and \gls{threshold} (b) generality metrics on the deoxyfluorination dataset.
    }
    \label{fig:add_results_task-selection_deoxyfluorination}
\end{figure*}

As outlined in the main text, we evaluate different substrate-specific acquisition functions: (1) \gls{pvfunc} and (2) \gls{rafunc}.
We find differing effects for the \gls{seqonela} and \gls{seqtwola} acquisition strategies: while the former is not significantly influenced by substrate acquisition function, the latter shows higher \gls{gap} values at later optimization stages for the \gls{pvfunc}.
The former observation is consistent across all datasets for both generality metrics.
The latter observation is again mainly driven by the borylation and deoxyfluorination datasets (see \cref{fig:add_results_task-selection_borylation,fig:add_results_task-selection_deoxyfluorination}, respectively), that were previously identified as harder optimizations (see above).
This observation shows that posterior variance-based substrate selection is important for difficult generality-oriented optimizations, where minimizing model uncertainty over the more complicated joint optimization space increases optimization efficiency.

\begin{figure*}[t]
    \centering
    \includegraphics[width=\linewidth]{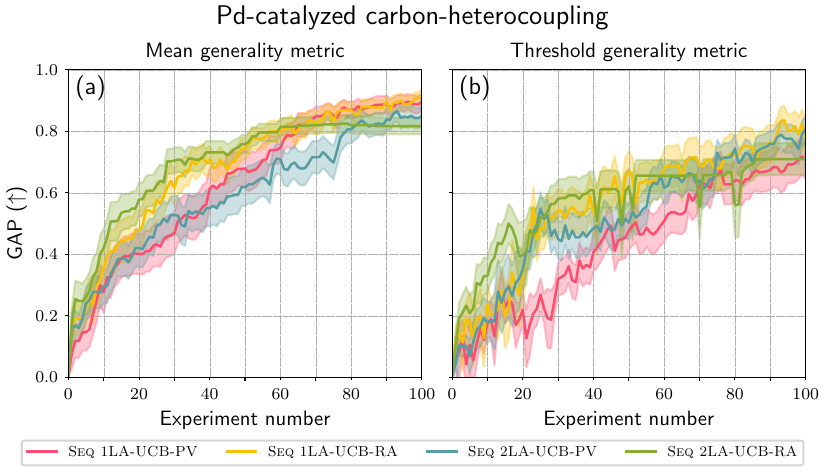}
    \vspace{-1em}
    \caption{
        Optimization trajectories for the \gls{seqonela} and \gls{seqtwola} acquisition strategies with \gls{pvfunc} and \gls{rafunc} substrate acquisition functions, on the \gls{mean} (a) and \gls{threshold} (b) generality metrics on the Pd-catalyzed carbon-heterocoupling dataset.
    }
    \label{fig:add_results_task-selection_Cernak}
\end{figure*}
\begin{figure*}[t]
    \centering
    \includegraphics[width=\linewidth]{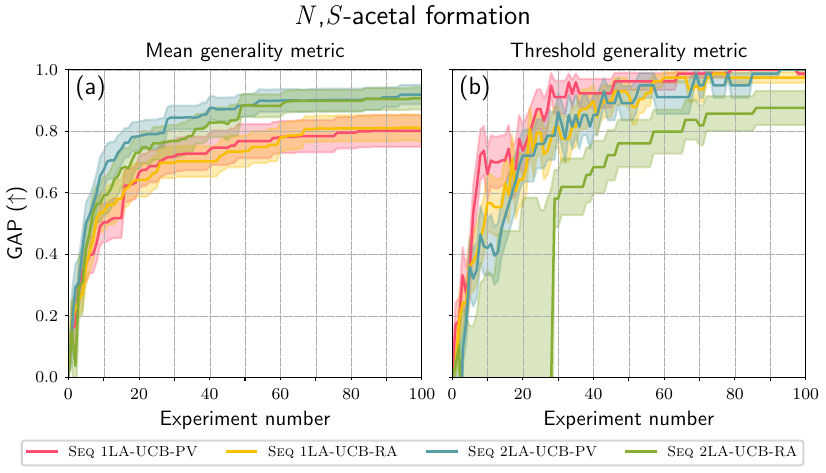}
    \vspace{-1em}
    \caption{
        Optimization trajectories for the \gls{seqonela} and \gls{seqtwola} acquisition strategies with \gls{pvfunc} and \gls{rafunc} substrate acquisition functions, on the \gls{mean} (a) and \gls{threshold} (b) generality metrics on the \textit{N},\textit{S}-acetal formation dataset.
    }
    \label{fig:add_results_task-selection_Denmark}
\end{figure*}

\textbf{Influence of Conditions Acquisition Function}

\begin{figure*}[t]
    \centering
    \includegraphics[width=\linewidth]{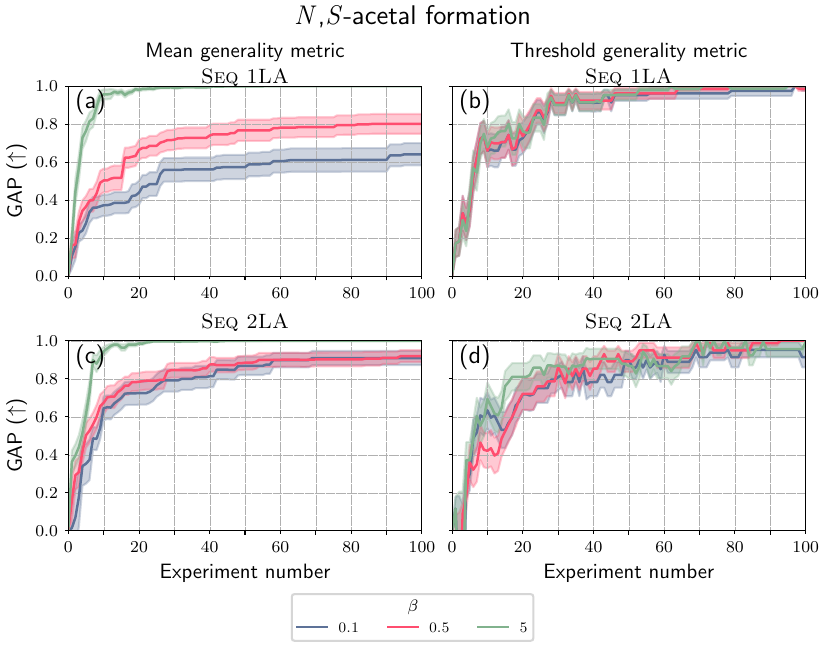}
    \vspace{-1em}
    \caption{
        Optimization trajectories for the \gls{seqonela} and \gls{seqtwola} acquisition strategies with varying exploration/exploitation of the conditions acquisition function, on the \gls{mean} (left) and \gls{threshold} (right) generality metrics on the \textit{N},\textit{S}-acetal formation dataset.
    }
    \label{fig:add_results_parameter-selection_Denmark}
\end{figure*}

In the main text, we compare different acquisition functions for conditions selection, identifying a highly explorative acquisition driver as a key driver for efficient optimization.
Investigating this relation across the different datasets, we find that the biggest effect is obtained on the \textit{N},\textit{S}-acetal formation dataset (see \cref{fig:add_results_parameter-selection_Denmark}).
However, across all datasets and generality metrics, \gls{seqtwolaucbfivepv} outperforms or matched their less explorative counterpart \gls{seqonelaucbzerofivepv}.
While we cannot identify trends when increased efficiency is observed, we note that the trend holds that explorative condition selection increases the likelihood of an efficient generality-oriented optimization.

\begin{figure*}[t]
    \centering
    \includegraphics[width=\linewidth]{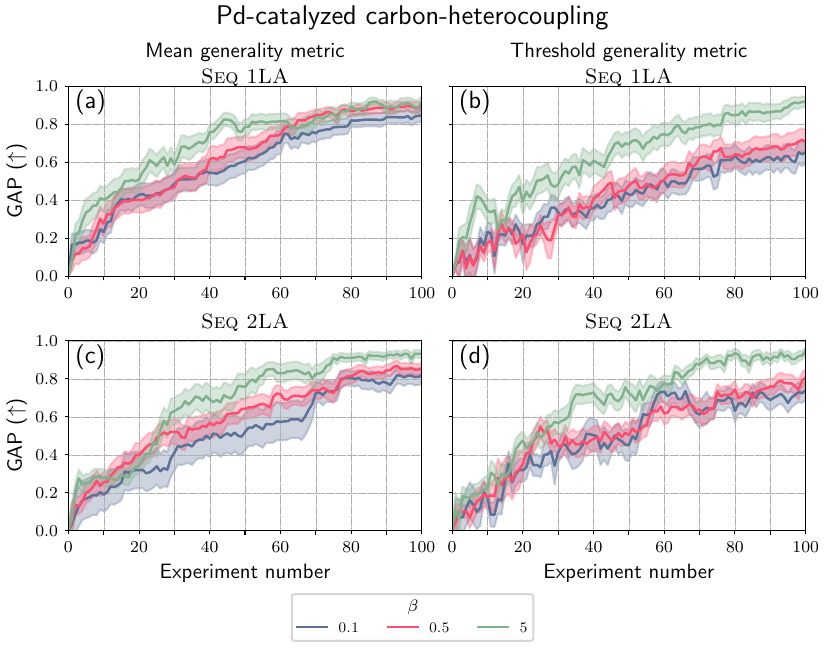}
    \vspace{-1em}
    \caption{
        Optimization trajectories for the \gls{seqonela} and \gls{seqtwola} acquisition strategies with varying exploration/exploitation of the conditions acquisition function, on the \gls{mean} (left) and \gls{threshold} (right) generality metrics on the Pd-catalyzed carbon-heterocoupling dataset.
    }
    \label{fig:add_results_parameter-selection_Cernak}
\end{figure*}
\begin{figure*}[t]
    \centering
    \includegraphics[width=\linewidth]{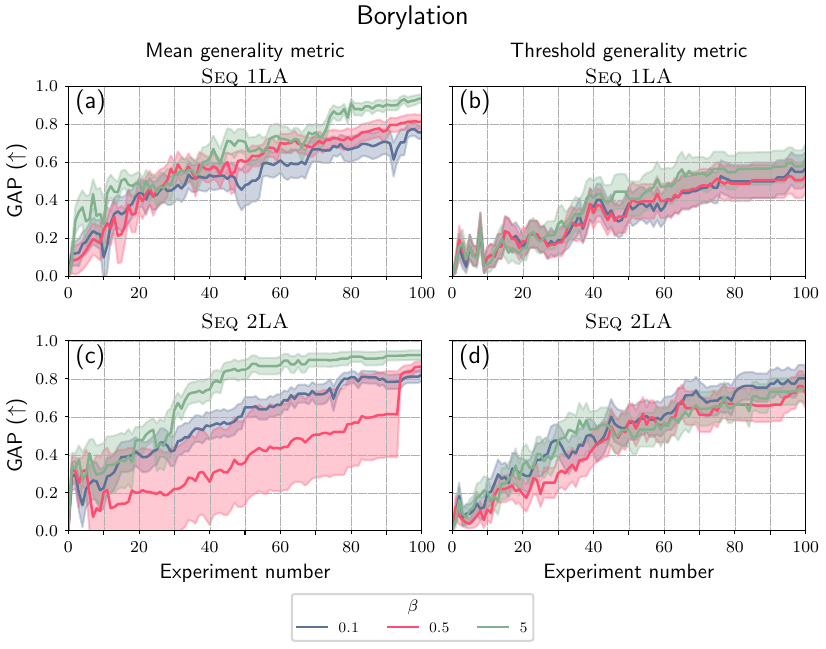}
    \vspace{-1em}
    \caption{
        Optimization trajectories for the \gls{seqonela} and \gls{seqtwola} acquisition strategies with varying exploration/exploitation of the conditions acquisition function, on the \gls{mean} (left) and \gls{threshold} (right) generality metrics on the borylation dataset. \sx{just a personal preference but are these individual figures meant to be centered in the middle of the page versus align to the top?}
    }
    \label{fig:add_results_parameter-selection_Borylation}
\end{figure*}
\begin{figure*}[t]
    \centering
    \includegraphics[width=\linewidth]{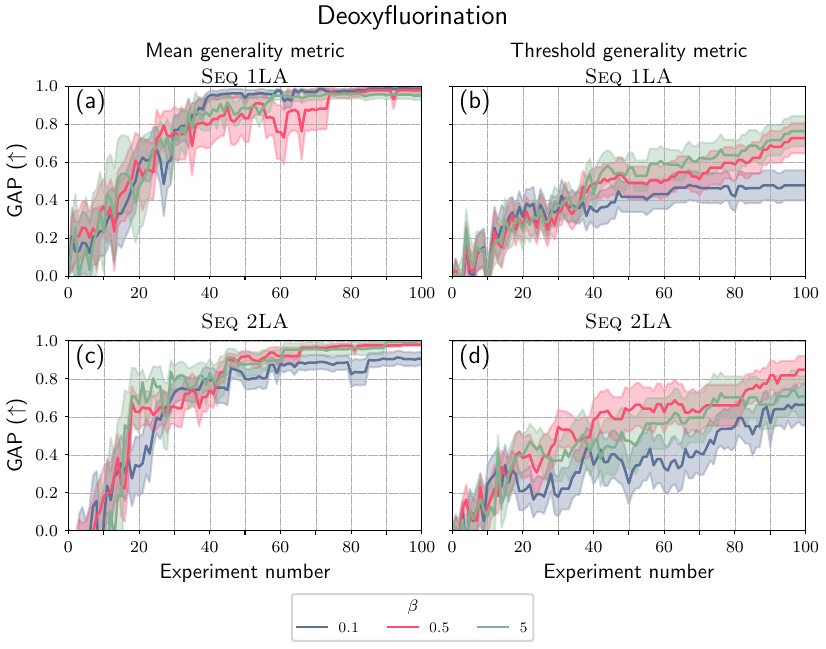}
    \vspace{-1em}
    \caption{
        Optimization trajectories for the \gls{seqonela} and \gls{seqtwola} acquisition strategies with varying exploration/exploitation of the conditions acquisition function, on the \gls{mean} (left) and \gls{threshold} (right) generality metrics on the deoxyfluorination dataset.
    }
    \label{fig:add_results_parameter-selection_Deoxyfluorination}
\end{figure*}

\textbf{Comparing Insights-Derived Strategy with State-of-the-Art}

In the main text, we used the obtained insights on conditions and substrate acquisition to design an acquisition policy for the next conditions and substrate that beats the current state-of-the-art for general conditions optimization.
Investigating the performance on each dataset, we find that the efficiency increase can be observed across all datasets, particularly for the \gls{mean} generality metric.
In contrast, for the \gls{threshold} generality metric, \gls{seqonelapipv} \citep{angello_closed-loop_2022} achieves significantly higher \gls{gap} values than for the \gls{mean} metric, showing on par performance to \gls{seqtwolaucbfivepv} in early optimization stages on three of the four datasets.
The surprisingly low optimization efficiency can be explained by an underexploration of the condition space (see \cref{tab:add_results_unique_conditions}): the \gls{pifunc} condition acquisition function essentially estimates which conditions have the highest probability of surpassing the maximum.
As the maximum is also estimated from the model due to the partial monitoring, the probability is deemed highest for the conditions that already are the maximum, which can be evaluated for $N$-many substrates, before different conditions have to be acquired.
Notably, \gls{seqonelapipv} explores more conditions for the \gls{threshold} generality metric, for which it also achieves higher \gls{gap} values in our experiments.

\begin{table}[t]
\centering
\caption{Average and standard deviation of unique conditions explored for \gls{seqtwolaucbfivepv} and \gls{seqonelapipv} for the two generality metrics and four datasets.}
\label{tab:add_results_unique_conditions}

\begin{tabularx}{\linewidth}{p{3cm} *{4}{>{\centering\arraybackslash}X}}
& \multicolumn{2}{c}{\gls{mean} generality metric} 
& \multicolumn{2}{c}{\gls{threshold} generality metric} \\
\cmidrule(lr){2-3} \cmidrule(lr){4-5}
& \gls{seqtwolaucbfivepv} & \gls{seqonelapipv} & \gls{seqtwolaucbfivepv} & \gls{seqonelapipv} \\
\midrule
Pd-catalyzed carbon-heterocoupling & $56.1\pm6.3$ & $3.9\pm2.2$ & $47.3\pm9.1$ & $10.7\pm6.4$ \\
\textit{N},\textit{S}-acetal formation & $33.3\pm3.3$ & $2.0\pm0.4$ & $25.6\pm9.0$ & $5.1\pm2.4$  \\
Borylation & $42.5\pm3.0$ & $3.2\pm1.2$ & $26.9\pm5.9$ & $11.0\pm6.6$ \\
Deoxyfluorination & $19.4\pm1.0$ & $2.8\pm1.1$ & $16.8\pm1.8$ & $6.2\pm2.8$ \\
\end{tabularx}

\end{table}

Across all datasets, \gls{random} does not obtain a \gls{gap} value significantly different from $0$, underlining that it is randomly guessing the most general conditions at each step.
Due to its initialization phase outlined above, \gls{bandit} is also randomly guessing the most general conditions until each condition has been tested at least once.
This behaviour can be nicely seen by the fact that the \gls{bandit} strategy only achieves \gls{gap} values significantly different from $0$ once the experiment number is equal to the number of possible condition combinations (\textit{e.g.}, $46$ for the \textit{N},\textit{S}-acetal formation dataset, as described above).

\begin{figure*}[t]
    \centering
    \includegraphics[width=\linewidth]{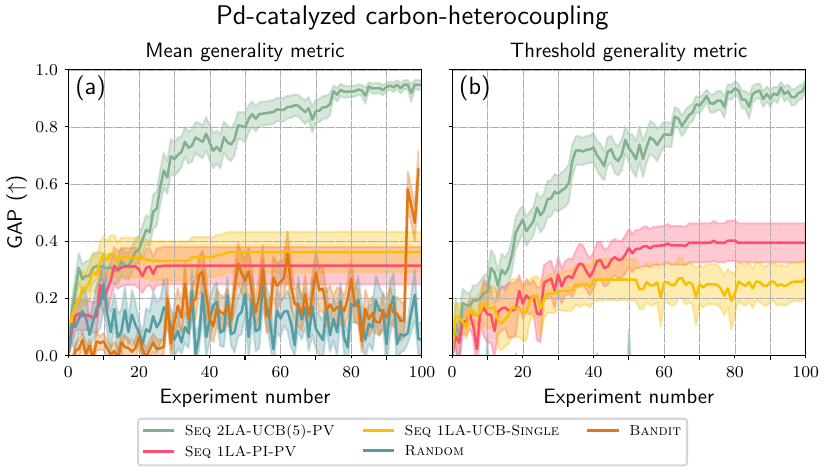}
    \vspace{-1em}
    \caption{
        Optimization trajectories of different algorithms for generality-oriented optimization previously reported in the chemical domain on the Pd-catalyzed carbon-heterocoupling dataset. Note that the \textsc{Bandit} algorithm is incompatible with the threshold aggregation function.
    }
    \label{fig:add_results_chemistry-algos_Cernak}
\end{figure*}
\begin{figure*}[t]
    \centering
    \includegraphics[width=\linewidth]{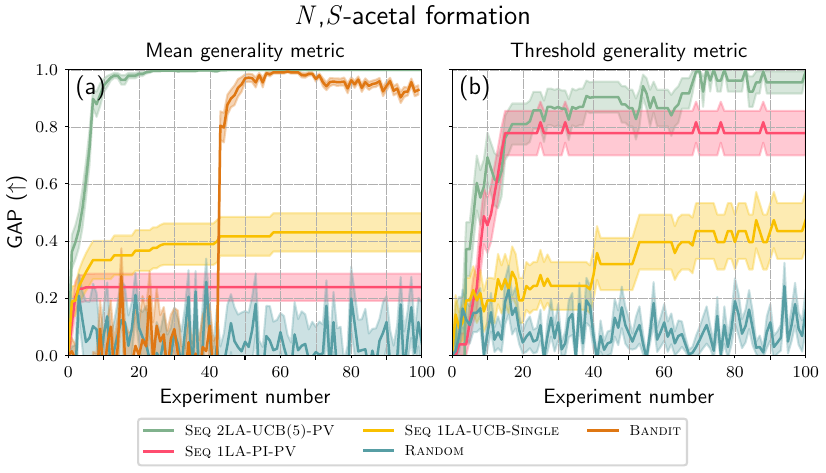}
    \vspace{-1em}
    \caption{
        Optimization trajectories of different algorithms for generality-oriented optimization previously reported in the chemical domain on the \textit{N},\textit{S}-acetal formation dataset. Note that the \textsc{Bandit} algorithm is incompatible with the threshold aggregation function.
    }
    \label{fig:add_results_chemistry-algos_Denmark}
\end{figure*}
\begin{figure*}[t]
    \centering
    \includegraphics[width=\linewidth]{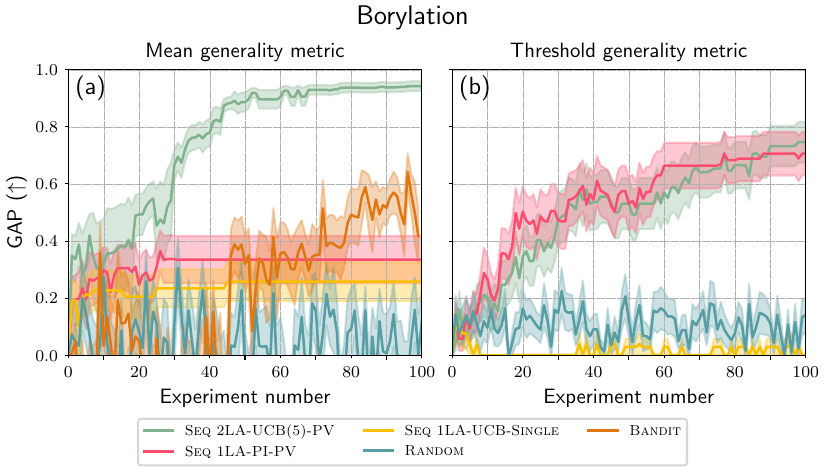}
    \vspace{-1em}
    \caption{
        Optimization trajectories of different algorithms for generality-oriented optimization previously reported in the chemical domain on the borylation dataset. Note that the \textsc{Bandit} algorithm is incompatible with the threshold aggregation function.
    }
    \label{fig:add_results_chemistry-algos_Borylation}
\end{figure*}
\begin{figure*}[t]
    \centering
    \includegraphics[width=\linewidth]{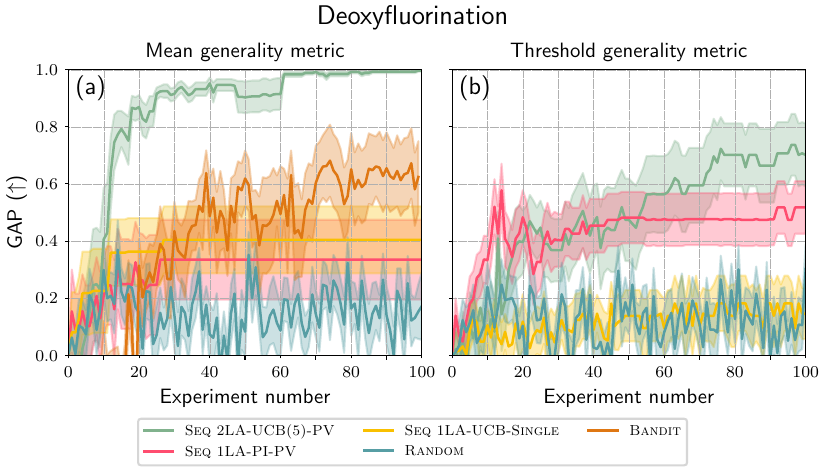}
    \vspace{-1em}
    \caption{
        Optimization trajectories of different algorithms for generality-oriented optimization previously reported in the chemical domain on the deoxyfluorination dataset. Note that the \textsc{Bandit} algorithm is incompatible with the threshold aggregation function.
    }
    \label{fig:add_results_chemistry-algos_Deoxyfluorination}
\end{figure*}

\end{document}